\documentclass{article}

\PassOptionsToPackage{numbers}{natbib}
 \usepackage[preprint]{neurips_2026}
 \usepackage{lineno}

\usepackage[utf8]{inputenc} 
\usepackage[T1]{fontenc}    
\usepackage{hyperref}       
\usepackage{url}            
\usepackage{booktabs}       
\usepackage{amsfonts}       
\usepackage{nicefrac}       
\usepackage{microtype}      
\usepackage{xcolor}         
\definecolor{varcolor}{RGB}{60,130,120}
\definecolor{syntaxcolor}{RGB}{180,40,40}
\definecolor{algcolor}{RGB}{40,80,180}

\usepackage{multirow}
\usepackage{graphicx}
\usepackage{amsmath}
\usepackage{cleveref}
\usepackage{algorithm}
\usepackage{algorithmic}
\usepackage{enumitem}
\usepackage{subcaption}

\usepackage{listings}

\makeatletter
\AtBeginDocument{
  \let\c@lstlisting\c@figure

  \renewcommand*{\ext@lstlisting}{lof}
}
\makeatother

\crefname{lstlisting}{figure}{figure}
\Crefname{lstlisting}{Figure}{Figure}

\crefname{section}{Appendix}{Appendices}
\Crefname{section}{Appendix}{Appendices}    

\title{Automated Kernel Discovery Towards Understanding \\ High-dimensional Bayesian Optimization}

\author{%
  Taeyoung Yun$^{1,*}$\quad Woocheol Shin$^{1,*}$\quad Inhyuck Song$^1$\quad Jaewoo Lee$^1$\quad Jinkyoo Park$^{1}$ \\
    $^1$Korea Advanced Institute of Science and Technology (KAIST)\\
    \texttt{\{99yty, woofe, son9ih, jaewoo, jinkyoo.park\}@kaist.ac.kr}
}

\begin{document}

\maketitle

\begin{abstract}
  Gaussian Process (GP) kernels are central to Bayesian optimization (BO), yet designing effective kernels for high-dimensional problems still relies on extensive manual engineering. Existing automated approaches struggle in high dimensions for two bottlenecks: their kernel search space is limited to additions and multiplications of base kernels, and LLM-based approaches require conditioning on raw observations, which becomes infeasible due to context-length limits and the difficulty of extracting meaningful patterns. We introduce \textbf{Kernel Discovery}, a LLM-driven evolutionary framework for high-dimensional BO that searches a broader kernel space beyond predefined composition rules and does not require conditioning on observations. Motivated by the observation that directly prompting an LLM to generate kernel code yields syntactically varied but functionally identical kernels, we adopt a two-stage approach: an LLM first proposes novel mathematical forms, then a second LLM call converts each form into validated, executable code. We also propose a leave-one-out continuous ranked probability score (LOO-CRPS) as a selection criterion that penalizes overfitted kernels. On five high-dimensional BO benchmarks, our method achieves an average rank of \textbf{1.2 out of 17}, outperforming competitive baselines. 
  We further analyze the discovered kernels to identify which kernels lead to improvements in high-dimensional BO.

\end{abstract}

\section{Introduction}
Optimizing high-dimensional black-box functions spans a wide range of machine learning problems, including hyperparameter optimization \citep{turner2021bayesian, wang2022pre}, neural architecture search \citep{swersky2014raiders,kandasamy2018neural}, biological sequence design \citep{maus2022local,leelatent}, and control tasks \citep{wang2020learning, berkenkamp2023bayesian}. Bayesian optimization (BO) is the de facto paradigm for black-box optimization, which iteratively selects promising candidates based on surrogate models fitted to observations so far \citep{kushner1964new,garnett2023bayesian}. However, it often suffers from the curse of dimensionality, as large distances between observations make accurate surrogate modeling difficult and exacerbate overexploration toward the boundary of the search space.

Prior works in high-dimensional BO have exploited structural assumptions such as additive decompositions \citep{duvenaud2011additive,kandasamy2015high,gardner2017discovering,lu2022additive}, low-dimensional subspaces \citep{wang2016bayesian,nayebi2019framework,letham2020re}, or sparsity \citep{eriksson2021high,papenmeier2022increasing}. However, these approaches perform well only on synthetic benchmarks with explicit underlying structures, which rarely hold in practice. Another family of methods employs trust regions to effectively search high-dimensional spaces \citep{wang2020learning,eriksson2019scalable}. While locality achieves promising results across diverse tasks \citep{maus2022local,eriksson2021scalable,maus2024joint}, it requires a large number of evaluations and may struggle to escape local optima.

Recently, \citet{hvarfner2024vanilla} and \citet{xu2024standard} showed that even with basic kernels, such as RBF or Matérn52 kernels, we can match or exceed specialized methods in high-dimensional BO by proper scaling of the lengthscale prior with dimensionality. Moreover, \citet{oh2018bock} and \citet{doumontwe} demonstrated that geometric input warpings (e.g., hypercylinder or hypersphere) with simple models can surpass sophisticated baselines on high-dimensional, real-world benchmarks. Although these breakthroughs are promising, identifying such effective priors or transformations still requires extensive manual engineering by domain experts. This suggests that automating kernel design with a system capable of mathematical reasoning could substantially reduce this barrier.

\begin{figure}[t]  
    \centering
    \includegraphics[width=.95\linewidth]{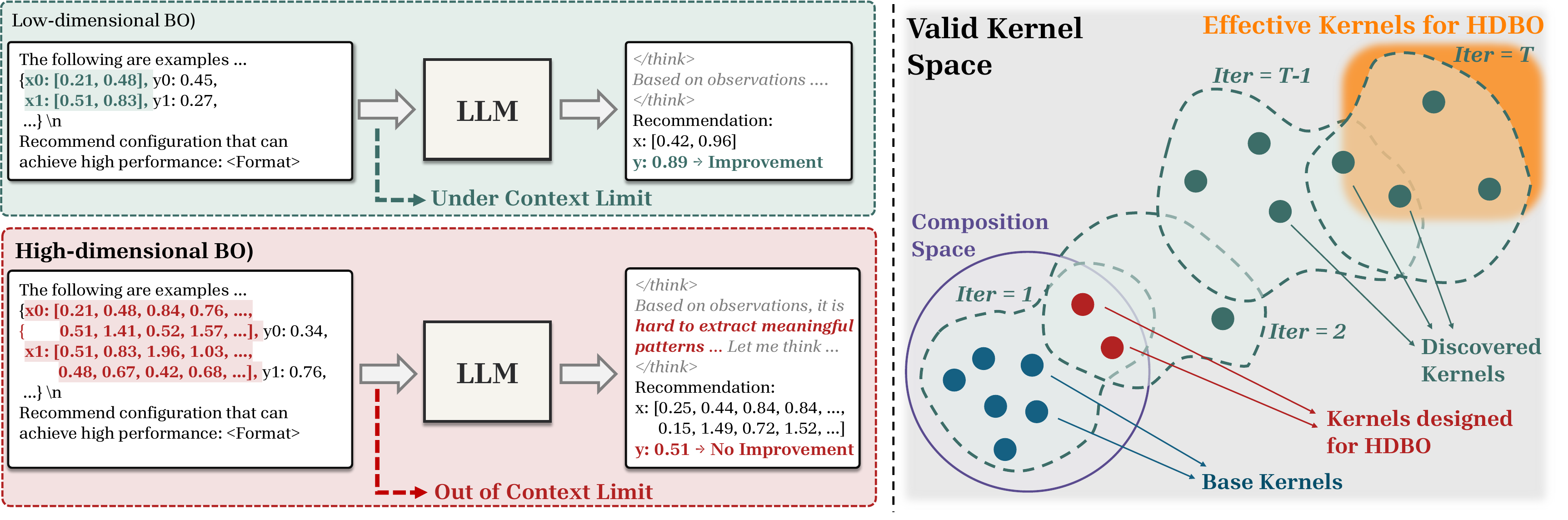}  
    \vspace{-5pt}
    \caption{Motivation Figure. \textbf{(Left):} In high-dimensional BO, a conventional LLM-based BO does not work due to the out-of-context limits and the difficulty of extracting meaningful patterns. \textbf{(Right):} Our pipeline enables us to discover effective kernels for high-dimensional BO beyond compositions.}
    \label{fig:motivation}
    \vspace{-15pt}
\end{figure}

This naturally prompts us to leverage Large Language Models (LLMs, \citealp{grattafiori2024llama,yang2025qwen3,comanici2025gemini}), which have internalized vast mathematical knowledge and can propose novel functional forms when incorporated into an evolutionary algorithm (EA) pipeline \citep{romera2024mathematical,novikov2025alphaevolve,aglietti2025funbo,zhao2026trajevo,kim2026self}. However, existing LLM-based BO methods do not directly scale to high-dimensional problems due to two bottlenecks. First, an \textbf{expressive bottleneck}: Existing kernel search methods are limited to additive and multiplicative compositions of base kernels, thereby limiting their ability to discover novel kernels suitable for high-dimensional BO. Second, an \textbf{interface bottleneck}: most LLM-based BO methods require conditioning on raw observations, which becomes infeasible in high dimensions due to context-length limits and the difficulty of extracting patterns from a long stream of numbers, as shown in \Cref{fig:motivation}.


To address these limitations, we introduce \textbf{Kernel Discovery}, a novel framework for designing effective kernel structures for high-dimensional BO.
First, we \textbf{initialize} the population with kernels tailored for high-dimensional BO. At each BO iteration, we \textbf{discover} new kernels through a two-stage approach: an LLM first proposes novel mathematical forms, and a second LLM converts the form into executable code. 
Through this decomposition, we encourage LLMs to explore the broader space of valid
kernels by leveraging mathematical reasoning capabilities. 
We then \textbf{validate} the discovered kernels via agnostic execution and PSD checks, retaining only valid GP kernels.
To \textbf{choose} the most promising kernel, we introduce LOO-CRPS, which is less prone to in-sample overfitting than the marginal log-likelihood metric. Finally, we evaluate the objective at the selected point and \textbf{update} the dataset and population, repeating until the evaluation budget is exhausted.

We conduct experiments on five high-dimensional BO benchmarks and achieve an average rank of \textbf{1.2 out of 17}, outperforming competitive baselines. We also conduct several ablation studies on the design choices of our framework. Finally, we analyze the discovered kernels and uncover that unexpected kernels (e.g., compositions of geometric warping or non-stationary components) might lead to improvements, providing  insights into what makes a kernel effective in high dimensions.

\section{Related Work}
\paragraph{High-dimensional Bayesian Optimization.} Several approaches have been proposed to push the limit of BO in high dimensions. One line of work exploits explicit structural assumptions, such as low-dimensional subspaces \citep{wang2016bayesian,nayebi2019framework,letham2020re,kirschner2019adaptive}, sparsity \citep{eriksson2021high,papenmeier2022increasing}, variable selection \citep{wang2020learning,song2022monte,hellsten2023high}, and additive decompositions \citep{duvenaud2011additive,kandasamy2015high,gardner2017discovering,lu2022additive}. However, such assumptions rarely hold in real-world problems. Another line of work employs trust regions, in which we constrain the search space to prevent the search among boundary points in high-dimensional space \citep{maus2022local,eriksson2019scalable,maus2024joint,daulton2022multi,maus2022discovering,ascia2025feasibility}. While it achieves superior performance across various domains, particularly in scalable settings, it requires too large a number of evaluations and remains susceptible to being trapped in local optima.

More recently, a series of surprising findings have challenged longstanding intuitions about high-dimensional BO. \citet{hvarfner2024vanilla} and \citet{xu2024standard} showed that simple kernels with well-chosen lengthscale priors can match or exceed sophisticated methods. \citet{papenmeier2025understanding} also claimed that scaling the initial lengthscale and random axis-aligned subspace perturbation sampling (RAASP, \citealp{rashidi2024cylindrical}) can mitigate the vanishing gradient issue in acquisition function optimization in high dimensions. \citet{doumontwe} further demonstrated that the smoothness of surrogate models is a critical factor, and that spherical input mappings with a simple linear basis suffice for competitive performance. Together, these results reveal that the community's prevailing assumptions about what drives high-dimensional BO may be incomplete. 
This underscores the need for a more in-depth investigation of kernel design in high-dimensional BO, which is precisely the gap our work aims to fill.

\paragraph{LLMs for Black-box Optimization.} Recent advancements in LLMs have demonstrated strong capabilities for complex reasoning \citep{grattafiori2024llama,yang2025qwen3,comanici2025gemini}, prompting several works to integrate them into the BO pipeline. \citet{aglietti2025funbo} and \citet{ngoadaptive} utilize LLM to discover novel acquisition functions, while \citet{li2025llamea} and \citet{liularge} replace the entire BO loop with an LLM-driven system. However, these methods are primarily evaluated in low-dimensional settings and rely on prepending prior observations as context, which becomes infeasible in high dimensions due to context-length constraints. Furthermore, extracting meaningful structure from dense observations can be challenging even for recent, closed-source LLMs.

Most closely related to our work is CAKE \citep{suwandiadaptive}, which employs an LLM to select and combine base kernels to improve BO surrogates. While it shares our motivation for LLM-guided kernel design, it also relies on prior trials as context and restricts the search space to additive and multiplicative compositions of base kernels. In contrast, our framework does not condition on raw observations and actively explores a broader space of valid kernels beyond compositions, thereby enabling more expressive surrogate modeling for high-dimensional problems.

\begin{figure}[t]  
    \centering
    \includegraphics[width=.95\linewidth]{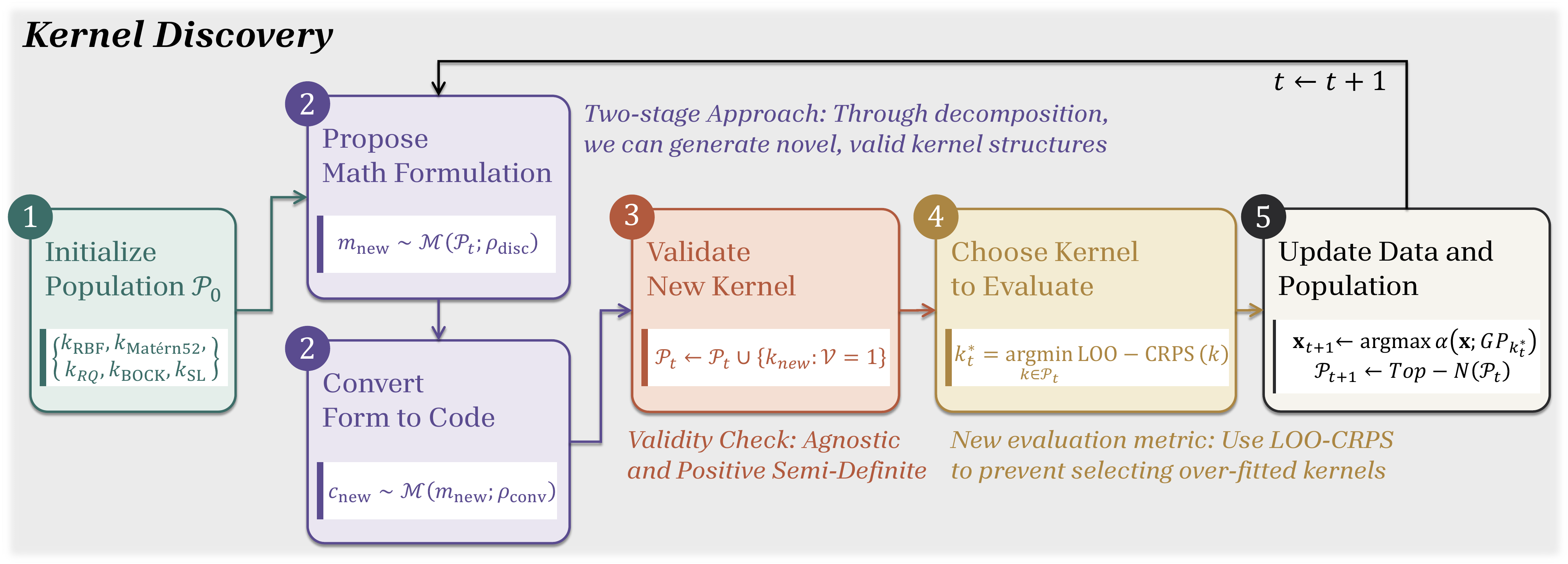}  
    \vspace{-5pt}
    \caption{Overview of the method. Given an initial population, we instruct an LLM to generate novel kernels via a two-stage approach: mathematical formulation and code conversion. We then perform a sanity check to retain only valid kernels and select the kernel with the lowest LOO-CRPS value. Finally, we propose a next query using the chosen kernel and update the dataset and population.}
    \vspace{-15pt}
    \label{fig:overview}
\end{figure}

\section{Preliminaries}
\paragraph{Bayesian Optimization.} In BO, we aim to find an input $\mathbf{x}\in\mathcal{X}$ that maximizes the unknown black-box function $f:\mathcal{X}\rightarrow \mathbb{R}$, 
which is typically expensive to evaluate and not differentiable. BO can be modularized into three orthogonal components: a surrogate model, an acquisition function, and an optimizer for the acquisition function \citep{tripp2026modularity}. At each iteration, we fit the surrogate model to the observations so far and select the next query that maximizes the chosen acquisition function using the optimizer. While there are several possible options for each component, the conventional approach is to use a Gaussian Process (GP, \citealp{rasmussen2003gaussian}) as a surrogate model, Expected Improvement (\texttt{EI}, \citealp{mockus1998application}) as an acquisition function, and \texttt{L-BFGS-B} \citep{byrd1995limited} as an optimizer. 

\paragraph{Gaussian Processes and Kernels.} A GP defines a distribution over functions and is specified by a mean function (constant in most cases) and a covariance matrix $\mathbf{K}$. The covariance matrix can be derived by using a kernel function $k(\mathbf{x}, \mathbf{x}'):\mathcal{X}\times\mathcal{X}\rightarrow\mathbb{R}$. The most frequently used kernel for BO is the RBF kernel \citep{williams1995gaussian}, which can be defined as follows:
\begin{equation}
    k_{\text{RBF}}(\mathbf{x}, \mathbf{x}') := \sigma^2 \exp\left(-\frac{1}{2}(\mathbf{x} - \mathbf{x}')^\top \mathbf{L}^{-1} (\mathbf{x} - \mathbf{x}')\right)
\end{equation}
where $\sigma^2 > 0$ is the output scale and $\mathbf{L} = \text{diag}(\ell_1^2, \ldots, \ell_d^2)$ is a diagonal matrix of per-dimension lengthscale hyperparameters $\ell_i > 0$.

Kernels can be combined or constructed in several ways. First, the sum and product of two valid kernels are also valid kernels, enabling the construction of richer covariance structures. Second, for any feature map $\phi:\mathcal{X}\rightarrow\mathbb{R}^m$, the inner product $k(\mathbf{x}, \mathbf{x}') = \phi(\mathbf{x})^\top \phi(\mathbf{x}')$ defines a positive semi-definite kernel by construction. By injecting these compositional and feature-map properties as inductive biases into an LLM-driven kernel construction process, we can discover novel kernel structures tailored for high-dimensional BO.

\section{Methods}
In this section, we introduce \textbf{Kernel Discovery}, a novel framework for discovering effective kernels for high-dimensional BO by leveraging LLMs as an evolutionary operator. We first initialize the population with well-known base kernels tailored for high dimensions. 
Then, we guide the LLM to generate novel kernel structures as mathematical forms and convert them into executable Python code. We conduct a sanity check to determine whether the generated kernel is valid. We also introduce a new selection criterion to choose the most promising kernel for acquisition function optimization. We repeat the process until convergence. We summarize the overview of our framework in \Cref{fig:overview}.

\vspace{3pt}
\textbf{Notations.} We represent each kernel $k=(m, c)$, where $m$ is its mathematical expression and $c$ is the corresponding code. Let $\mathcal{P}_t = \{k_1^{(t)}, \ldots, k_N^{(t)}\}$ denote the population of size $N$ at iteration $t$. We denote by $\mathcal{M}(\cdot\,;\,\rho)$ a call to the LLM conditioned on a role-specific prompt $\rho$. We also introduce an auxiliary verifier $\mathcal{V}$ to check whether the discovered kernel is valid.

\begin{figure}[t]  
    \centering
    \includegraphics[width=0.95\linewidth]{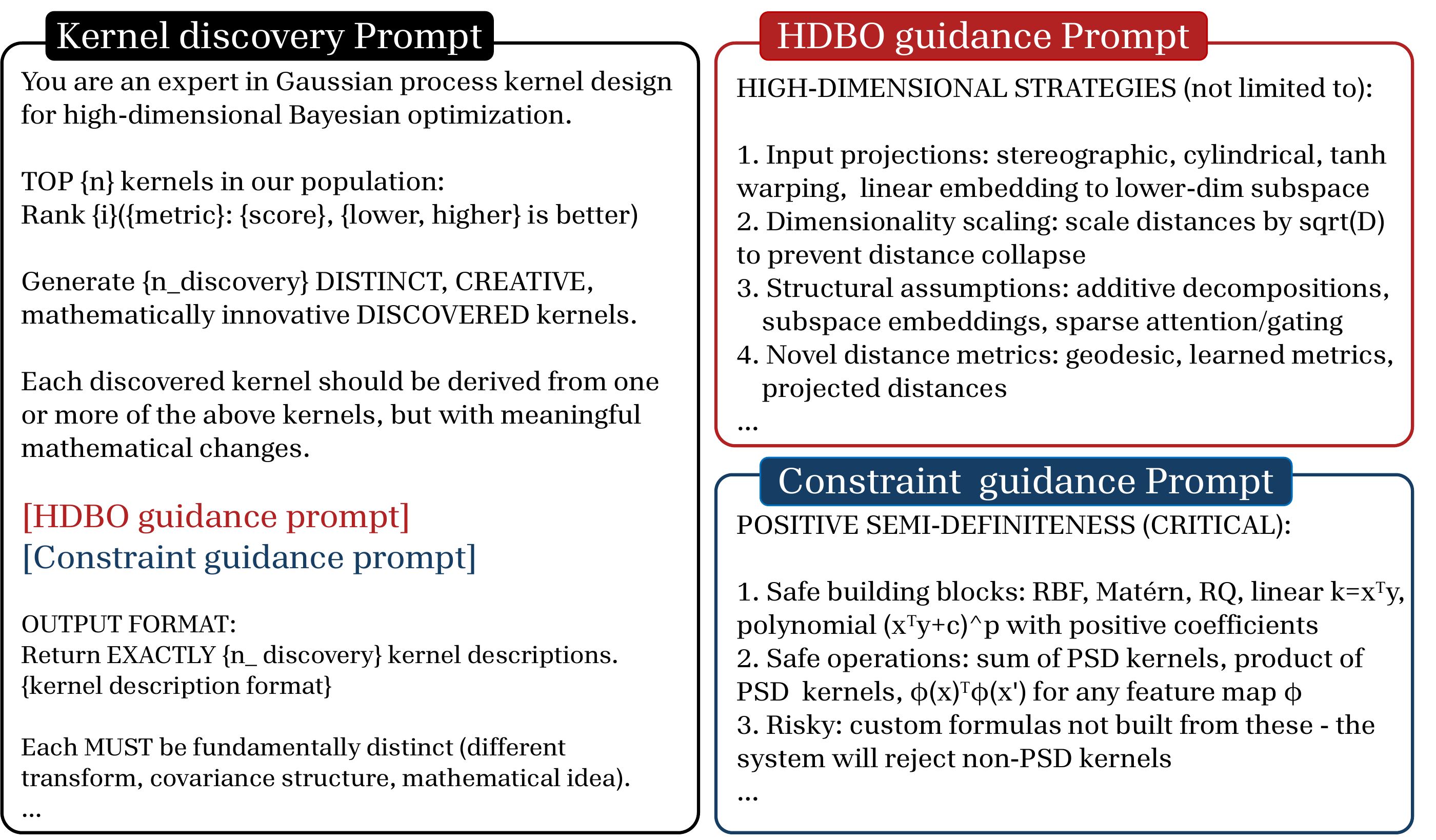}  
    \vspace{-5pt}
    \caption{
    Prompts for Kernel Discovery. To discover novel kernels for high-dimensional BO, we start with a prompt that simulates the reasoning of a human expert. Then, we instruct the LLM to understand prior strategies for high-dimensional BO and the constraints it should satisfy.
    }
    \label{fig:KD_prompt}
    \vspace{-12pt}
\end{figure}

\vspace{3pt}
\textbf{Initialize Population.} Similar to several approaches that integrate LLMs into the evolutionary loop \citep{novikov2025alphaevolve}, we initialize the base population with existing kernels that are widely used for BO. As our focus is on high-dimensional BO, we need to carefully select base kernels that will encourage LLMs to extract useful knowledge from them and give insights into how to propose better kernels. To this end, we set the initial population as follows:
\begin{equation}\label{eq:init}
    \mathcal{P}_0 = \{k_{\text{RBF}},\, k_{\text{Matérn52}},\, k_{\text{RQ}},\, k_{\text{BOCK}},\, k_{\text{SL}}\},
\end{equation}
For the RBF, Matérn52, and RQ, we use the lengthscale prior $\ell_{i}\sim\mathcal{LN}(\sqrt{2}+\log(\sqrt{D}), \sqrt{3})$, proposed by \citet{hvarfner2024vanilla}. 
BOCK \citep{oh2018bock} and SL \citep{doumontwe} propose geometric mappings to avoid overexploration. We do not include Linear and Periodic as they perform poorly in high dimensions.

\vspace{3pt}
\textbf{Discover Kernels with LLMs.} Starting from the initial population $\mathcal{P}_0$, we leverage LLMs to propose new kernel structures. 
Specifically, our discovery pipeline consists of two stages: mathematical formulation and code generation.

While one can directly prompt an LLM to generate novel kernels in the code generation space, this is undesirable, 
as LLMs often produce functionally redundant kernel codes (e.g., by renaming variables or substituting \texttt{x-y} with \texttt{x.sub(y)}) rather than genuinely novel structures.
To mitigate this, we first express the existing kernels in mathematical form $\{m_i^{(t)}\}_{i=1}^{N}$ and prompt the LLM to generate new mathematical forms that satisfy the constraints and are suitable for high-dimensional BO:
\begin{equation}
    m_{\text{new}} \sim \mathcal{M}\!\left(\{m_i^{(t)}\}_{i=1}^{N};\, \rho_{\text{disc}}\right)
\end{equation}
where $\rho_{\text{disc}}$ is a prompt for discovery stage as depicted in \Cref{fig:KD_prompt}.
After discovering a new mathematical form, we introduce another LLM to convert the formulation into executable code:
\begin{equation}
    c_{\text{new}} \sim \mathcal{M}(m_{\text{new}};\, \rho_{\text{conv}})
\end{equation}
where $\rho_{\text{conv}}$ is a prompt for conversion stage as depicted in \Cref{fig:prompt_conv}.
Rather than conditioning on raw observations, we provide only the mathematical forms from previous iterations as context. This dramatically reduces the context length, allowing the LLM to focus on structural novelty rather than sifting through high-dimensional numerical data for patterns. 

Following prior works, we find that composing kernels is the simplest way to discover new valid kernels. We therefore add a composition stage in which the LLM takes the top-$k$ kernels as input and combines them via addition or multiplication. You can find the prompt for composition in \Cref{fig:prompt_comp}.

\textbf{Validate Discovered Kernels.} Given a new candidate $k_{\text{new}}=(m_{\text{new}}, c_{\text{new}})$, we filter it with two empirical checks: that the code runs correctly across varying input shapes ($\mathcal{V}_{\text{agn}}$) and that the Cholesky decomposition succeeds on random test inputs ($\mathcal{V}_{\text{psd}}$). Formally:
\begin{equation}
    \mathcal{V}(k_{\text{new}}) = \mathcal{V}_{\text{agn}}(k_{\text{new}}) \wedge \mathcal{V}_{\text{psd}}(k_{\text{new}})
\end{equation}
where $\mathcal{V}_{\text{agn}}$ tests the \texttt{forward} function on inputs of varying batch size and dimensionality, and $\mathcal{V}_{\text{psd}}$ checks for a successful Cholesky decomposition. Only candidates with $\mathcal{V}(k_{\text{new}}) = 1$ are added to the current population pool $\mathcal{P}_{t}$.
\begin{equation}
    \mathcal{P}_{t} \leftarrow \mathcal{P}_t \cup \{k^{\text{new}} : \mathcal{V}(k^{\text{new}}) = 1\}
\end{equation}

\vspace{3pt}
\textbf{Choose the Promising Kernel.} After generating new kernels, we need to select which kernel $k\in\mathcal{P}_{t}$ should be used for the acquisition function maximization. While the marginal log-likelihood (MLL) is a straightforward approach, it may prefer overly complex kernels that overfit the current observations and sacrifice predictive calibration. To this end, we introduce the Leave-one-out Continuous Ranked Probability Score (LOO-CRPS), a selection criterion that scores held-out predictive distributions and thereby penalizes overconfident fits more directly than MLL, defined as follows:
\begin{equation}
    \text{CRPS}\!\left(\mathcal{N}(\mu,\sigma^2),\, y\right) 
    = \sigma \left[ 
        \frac{y - \mu}{\sigma} \left( 2\Phi\!\left(\frac{y-\mu}{\sigma}\right) - 1 \right) 
        + 2\,\phi\!\left(\frac{y-\mu}{\sigma}\right) 
        - \frac{1}{\sqrt{\pi}}
    \right]
\end{equation}
where $\phi$ and $\Phi$ are the standard normal PDF and CDF, respectively. 
While CRPS is more robust than MLL, it can still overfit when evaluated on training data. To alleviate this, we aggregate CRPS over leave-one-out predictives across all $n$ observations:
\begin{equation}
    \begin{aligned}
        &\text{LOO-CRPS}(\mathbf{K}_k, \mathbf{y}) 
        = \frac{1}{n} \sum_{i=1}^{n} \text{CRPS}\!\left(\mathcal{N}(\mu_{-i},\, \sigma^2_{-i}),\, y_i\right),
        \quad\mu_{-i} = y_i - \frac{[\mathbf{K}_k^{-1}\mathbf{y}]_i}{[\mathbf{K}_k^{-1}]_{ii}}, 
        \: \sigma^2_{-i} = \frac{1}{[\mathbf{K}_k^{-1}]_{ii}}
    \end{aligned}
\end{equation}
where $\mathbf{K}_k$ is a kernel matrix derived from the kernel $k$. 
By leveraging the standard LOO identity for GP, the predictive mean $\mu_{-i}$ and variance $\sigma^2_{-i}$ can be computed analytically from a single matrix inversion. At each round, we select
\begin{equation}
    k^\star_t = \arg\min_{k \in \mathcal{P}_t} \text{LOO-CRPS}(\mathbf{K}_k, \mathbf{y})
\end{equation}
Finally, we use the selected kernel $k^\star_t$ and fit its hyperparameters (e.g., lengthscale, scale, variance), maximize the acquisition function $\alpha(\mathbf{x};\, \text{GP}_{k^\star_t})$ to obtain the next query point $\mathbf{x}_{t+1}$, evaluate the corresponding objective $y_{t+1} = f(\mathbf{x}_{t+1})$, and augment the dataset $\mathcal{D}_{t+1} \leftarrow \mathcal{D}_{t} \cup \{(\mathbf{x}_{t+1}, y_{t+1})\}$.

\vspace{3pt}
\textbf{Update the Population.} To maintain a fixed population size $N$ across rounds, we then truncate $\mathcal{P}_{t+1}$ by retaining the $N$ members with the lowest LOO-CRPS:
\begin{equation}
    \mathcal{P}_{t+1} \leftarrow \operatorname{Top\text{-}N}\!\left(\mathcal{P}_t;\, \text{LOO-CRPS}\right).
\end{equation}
This elitist update ensures that the population progressively favors kernels with more calibrated predictives while preventing unbounded growth. We additionally discard kernels that fail to improve performance over consecutive rounds. We provide full details of the algorithm in \Cref{alg:main}. 

\begin{table*}[t]
\centering
\caption{Performance comparison of various methods across standard benchmarks. $D$ denotes the dimensionality of the task. \textbf{\textcolor{blue}{Blue}} denotes the best entry in the column, and \textbf{\textcolor{violet}{Violet}} denotes the second best. Experiments are conducted with 4 random seeds.}
\label{tab:benchmark_results}
\vspace{2pt}
\renewcommand{\arraystretch}{1.2} 

\resizebox{\textwidth}{!}{
\begin{tabular}{l ccccc c}
\toprule
\multirow{2}{*}{\textbf{Method}} & \textbf{Rover $(\uparrow)$} & \textbf{Mopta08 $(\downarrow)$} & \textbf{Lasso-DNA $(\downarrow)$} & \textbf{SVM $(\downarrow)$} & \textbf{Humanoid $(\uparrow)$} & \textbf{Average} \\
& ($D=100$) & ($D=124$) & ($D=180$) & ($D=388$) & ($D=6392$) & \textbf{Rank} \\
\midrule

\multicolumn{7}{l}{\textbf{Base Kernels}} \\
\midrule
RBF & 3.552 $\pm$ 0.304 & 217.75 $\pm$ 2.33 & 0.291 $\pm$ 0.001 & \textbf{\textcolor{violet}{0.061 $\pm$ 0.004}} & 503.00 $\pm$ 67.82\phantom{0} & \phantom{0}6.6 / 17 \\
Matérn52 & 3.453 $\pm$ 0.366 & \textbf{\textcolor{blue}{215.77 $\pm$ 0.57}} & 0.292 $\pm$ 0.003 & 0.063 $\pm$ 0.003 & 509.79 $\pm$ 102.32 & \phantom{0}6.8 / 17 \\
Linear & 3.950 $\pm$ 0.441 & 274.67 $\pm$ 8.87 & 0.312 $\pm$ 0.004 & 0.226 $\pm$ 0.003 & 435.49 $\pm$ 42.00\phantom{0} & 13.0 / 17 \\
Periodic & 3.664 $\pm$ 0.231 & 309.10 $\pm$ 5.01 & 0.332 $\pm$ 0.005 & 0.226 $\pm$ 0.002 & 413.83 $\pm$ 71.30\phantom{0} & 15.0 / 17 \\
BOCK & 3.725 $\pm$ 0.593 & 225.36 $\pm$ 1.82 & 0.291 $\pm$ 0.003 & 0.068 $\pm$ 0.007 & 669.52 $\pm$ 50.15\phantom{0} & \phantom{0}6.1 / 17 \\
SL & 4.096 $\pm$ 0.473 & 246.78 $\pm$ 3.92 & 0.297 $\pm$ 0.001 & 0.112 $\pm$ 0.007 & 637.72 $\pm$ 37.86\phantom{0} & \phantom{0}8.2 / 17 \\
RQ & 3.858 $\pm$ 0.515 & 217.53 $\pm$ 4.25 & 0.294 $\pm$ 0.002 & 0.069 $\pm$ 0.017 & 600.80 $\pm$ 152.49 & \phantom{0}6.6 / 17 \\
\midrule

\multicolumn{7}{l}{\textbf{Search-based}} \\
\midrule
Greedy Search & 3.606 $\pm$ 0.270 & 225.36 $\pm$ 1.82 & 0.291 $\pm$ 0.003 & 0.066 $\pm$ 0.004 & 687.12 $\pm$ 119.52 & \phantom{0}6.1 / 17 \\
Compositional Search & 3.869 $\pm$ 0.531 & 218.76 $\pm$ 1.17 & \textbf{\textcolor{violet}{0.288 $\pm$ 0.003}} & 0.067 $\pm$ 0.007 & \textbf{\textcolor{violet}{700.58 $\pm$ 81.55}}\phantom{0} & \phantom{0}\textbf{\textcolor{violet}{4.4 / 17}} \\
CAKE & 3.412 $\pm$ 0.586 & \phantom{0}231.40 $\pm$ 15.67 & 0.300 $\pm$ 0.011 & 0.131 $\pm$ 0.036 & 667.49 $\pm$ 24.99\phantom{0} & \phantom{0}9.8 / 17 \\
\midrule

\multicolumn{7}{l}{\textbf{LLM-based}} \\
\midrule
ARM & 3.981 $\pm$ 0.378 & 238.17 $\pm$ 3.65 & 0.319 $\pm$ 0.003 & 0.228 $\pm$ 0.003 & 457.14 $\pm$ 44.96\phantom{0} & 12.0 / 17 \\
ATRBO & 3.054 $\pm$ 0.816 & 267.65 $\pm$ 4.63 & 0.328 $\pm$ 0.010 & 0.221 $\pm$ 0.016 & 428.80 $\pm$ 295.37 & 15.2 / 17 \\
TREvol & 3.398 $\pm$ 0.620 & 236.41 $\pm$ 3.96 & 0.301 $\pm$ 0.007 & 0.079 $\pm$ 0.007 & 469.39 $\pm$ 144.43 & 11.6 / 17 \\
TROpt & \textbf{\textcolor{violet}{4.114 $\pm$ 0.409}} & 234.23 $\pm$ 1.86 & 0.300 $\pm$ 0.006 & 0.110 $\pm$ 0.005 & 520.38 $\pm$ 65.42\phantom{0} & \phantom{0}7.8 / 17 \\
TRPareto & 3.872 $\pm$ 0.467 & 253.45 $\pm$ 4.73 & 0.296 $\pm$ 0.001 & 0.185 $\pm$ 0.008 & 509.90 $\pm$ 114.06 & 10.0 / 17 \\
LMABO & 3.953 $\pm$ 0.369 & 240.70 $\pm$ 5.24 & 0.304 $\pm$ 0.006 & 0.226 $\pm$ 0.001 & 390.23 $\pm$ 44.78\phantom{0} & 12.6 / 17 \\
\midrule

\textbf{Kernel Discovery (Ours)} & \textbf{\textcolor{blue}{4.353 $\pm$ 0.319}} & \textbf{\textcolor{violet}{216.81 $\pm$ 0.87}} & \textbf{\textcolor{blue}{0.286 $\pm$ 0.001}} & \textbf{\textcolor{blue}{0.056 $\pm$ 0.003}} & \textbf{\textcolor{blue}{762.78 $\pm$ 72.39}}\phantom{0} & \phantom{0}\textbf{\textcolor{blue}{1.2 / 17}} \\
\bottomrule
\end{tabular}
}
\end{table*}

\section{Experiments}
We evaluate our framework on five high-dimensional BO benchmarks and compare against 16 baselines, including base kernels, search-based methods, and LLM-based methods. 
Our code is \href{https://github.com/Shin-woocheol/Kernel_discovery}{here}.

\vspace{3pt}
\textbf{Benchmarks.}
Following prior works on high-dimensional BO \citep{hvarfner2024vanilla,doumontwe}, we evaluate our framework on five standard benchmarks with dimensionality ranging from $D=100$ to $D=6392$:
(i) \textbf{Rover} ($D=100$, \citealp{wang2018batched}), (ii) \textbf{Mopta08} ($D=124$, \citealp{eriksson2021high}), (iii) \textbf{Lasso-DNA} ($D=180$, \citealp{vsehic2022lassobench}), (iv) \textbf{SVM} ($D=388$, \citealp{eriksson2021high}), (v) \textbf{Humanoid} ($D=6392$, \citealp{wang2020learning}). We exclude \textbf{Ant} since the function can be easily exploited by random initialization. Please refer to \Cref{app:task_details} for the detailed description of each task.

\vspace{3pt}
\textbf{Baselines.}
We compare against 16 baselines across three categories: (1) BO methods that rely on a single fixed kernel, (2) search-based kernel methods, and (3) LLM-based BO methods. Please refer to \Cref{app:baseline_details} for the detailed description of each baseline.
\begin{itemize}[leftmargin=2em]
    \item \textbf{Base kernels}: BO methods that rely on a single kernel: RBF, Matérn52, Linear, Periodic, RQ, BOCK, and SL.
    \item \textbf{Search-based Methods}: Methods that search over kernel candidates, including Greedy Search \citep{roman2019experimental}, Compositional Search \citep{duvenaud2013structure}, and CAKE \citep{suwandiadaptive}. For a fair comparison, all search-based methods share the same initial population $\mathcal{P}_0$ defined in \Cref{eq:init}.
    \item \textbf{LLM-based Methods}: LLM-based BO methods, including end-to-end approaches (ARM, ATRBO, TREvol, TROpt, and TRPareto) from the LLAMEA-BO \citep{li2025llamea}, and LMABO \citep{ngoadaptive}, which adaptively selects the acquisition function.
\end{itemize}

\begin{figure}[t]  
    \centering
    \includegraphics[width=1\linewidth]{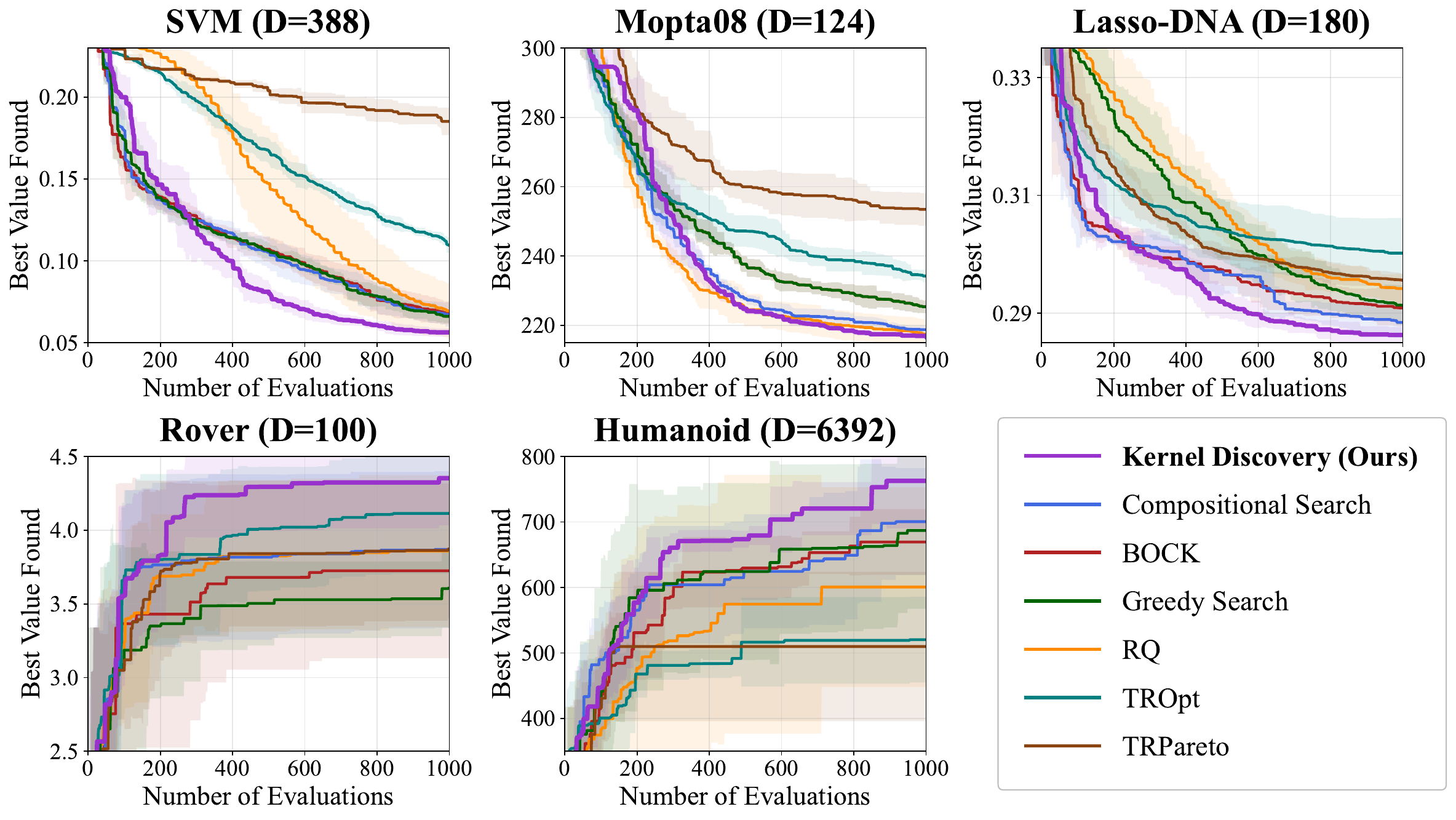}  
    \vspace{-15pt}
    \caption{Learning curve of various methods across standard benchmarks. We visualize the top-2 baselines by average rank per category for clarity. Please refer to \Cref{fig:main_extended} for the full results.}
    \label{fig:main}
    \vspace{-12pt}
\end{figure}

\vspace{3pt}
\textbf{Implementation.}
For all methods (except for LLM-based approaches that change other parts of the BO pipeline), we use \texttt{LogEI} \citep{ament2023unexpected} as an acquisition function and \texttt{L-BFGS-B} as an optimizer. We use the Sobol sequence \citep{sobol2011construction} to generate initial samples with size $\vert \mathcal{D}_{0}\vert=20$, use a batch size of $q=20$, and set the evaluation budget $T=1000$. For our method, we set the population size to $N=10$, generate two new kernel candidates (one from the discovery stage and one from the composition stage) per BO iteration. For a fair comparison, we use \texttt{GPT-4o} as the LLM backbone for both our method and all LLM-based baselines. Please refer to \Cref{app:implement_detail} for more details on implementation.


\vspace{3pt}
\textbf{Main Results.}
We summarize the results of all methods in \Cref{tab:benchmark_results}. 
As shown in the table, Kernel Discovery achieves an average rank of $\textbf{1.2}$ across benchmarks, the best among all $\textbf{17}$ methods. 
The runner-up is Compositional Search (rank $4.4$), confirming that actively searching over kernel candidates is beneficial. However, due to the restricted search space, it is insufficient to discover effective kernels in high dimensions. 
Notably, CAKE (rank $9.8$) performs substantially worse than competitive base kernels, indicating that injecting observations as context for the LLM is ineffective and degrades performance in high dimensions. 
Similarly, other LLM-based baselines underperform, indicating that leveraging LLMs to discover kernels within the BO pipeline is more effective than leveraging the LLM itself as a black-box optimizer.


We visualize the learning curve of kernel discovery against several baselines in \Cref{fig:main}. 
Our approach consistently improves throughout training by discovering novel kernel functions at each BO iteration. Compositional Search and BOCK sometimes perform well in early rounds but stagnate in suboptimal regions. Our method avoids this by continually evolving the kernel population. We also conduct a runtime analysis of our method and the baselines in \Cref{app:time_complexity}.

\vspace{-0.25cm}

\section{Ablation Studies}
\vspace{-0.25cm}

We discuss the key design choices in our pipeline, justifying the need for the two-stage approach and analyzing the importance of each component.

\begin{figure}[t]
    \centering
    \begin{subfigure}[b]{0.56\textwidth}
        \centering
        \includegraphics[width=1\linewidth]{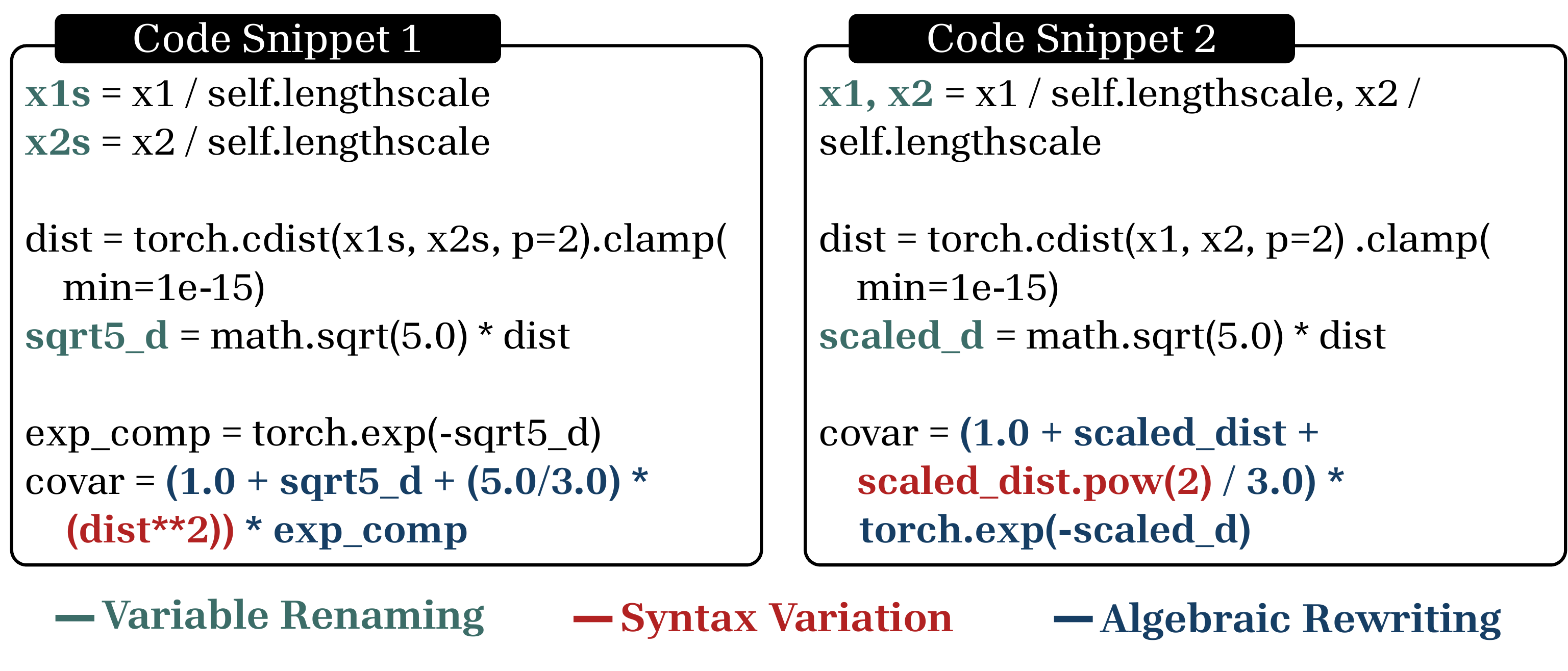}
        \caption{Example code snippets from direct code generation. Both snippets result in the same function.}
        \label{fig:direct_code_gen}
    \end{subfigure}
    \hfill 
    \begin{subfigure}[b]{0.4\textwidth}
        \centering
        \includegraphics[width=1\linewidth]{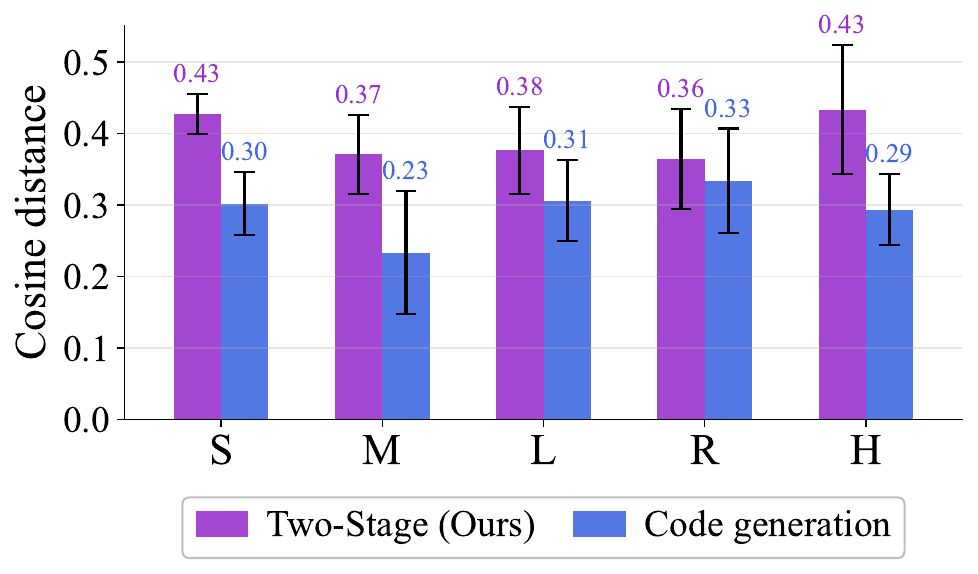}
        \caption{Cosine distance among kernels from direct code generation vs two-stage approach.}
        \label{fig:cosine_distance}
    \end{subfigure}
    \caption{Ablation studies on the two-stage approach for kernel discovery.}
    \vspace{-8pt}
\end{figure}

\textbf{Importance of Two-Stage Approach.}
We observed that when we directly instruct an LLM to generate code for novel kernel structures, it sometimes changes the variable names and replaces operators with equivalent operators, as shown in \Cref{fig:direct_code_gen}. 
We further demonstrate this phenomenon by measuring the functional diversity between generated kernels from direct code generation and those from our two-stage approach. As depicted in \Cref{fig:cosine_distance}, our approach consistently produces more genuinely diverse kernel candidates.
We also found that this functional diversity translates to consistently better performance (See \Cref{tab:cosine_distance_perf}), confirming that the two-stage approach discovers genuinely novel and effective kernel structures. Further details on the definition of the diversity metric, performance comparison, and functionally redundant code examples are provided in \Cref{app:direct_code_gen}.

\begin{figure}[t]
    \centering
    
    \begin{minipage}[c]{0.36\textwidth}
        \centering
        \includegraphics[width=1\linewidth]{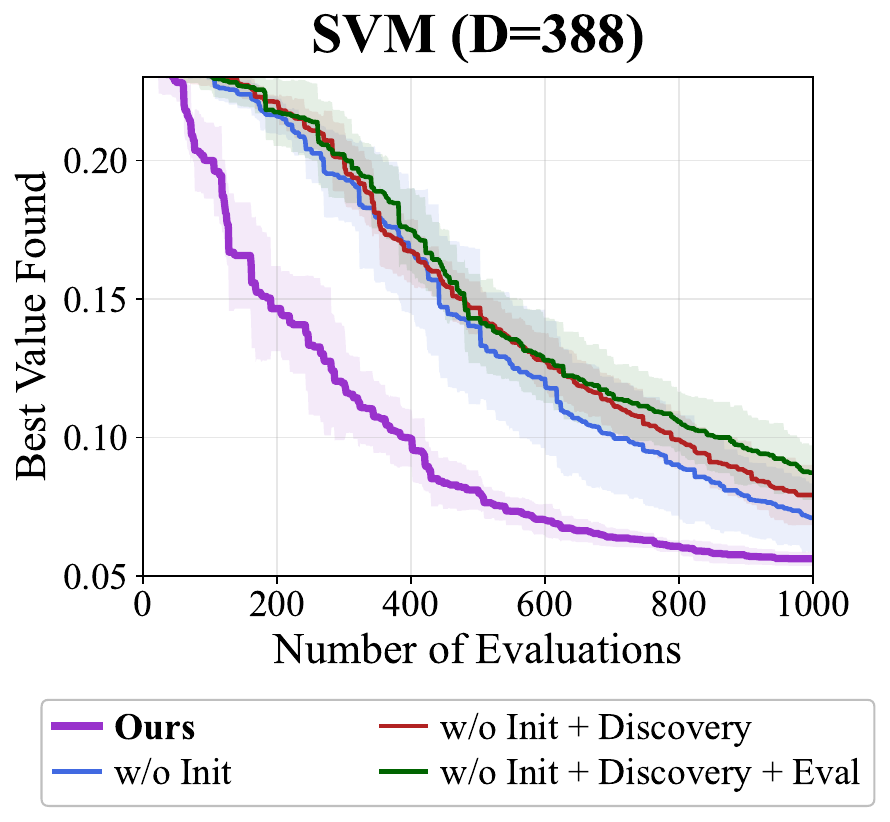}
        \captionof{figure}{Ablation on each component of Kernel Discovery.}
        \label{fig:system_arch}
    \end{minipage}
    \hfill
    \begin{minipage}[c]{0.6\textwidth}
        \centering
        
        \captionof{table}{Analysis on prompt template. Experiments are conducted with four random seeds in the SVM benchmark.}
        \label{tab:prompt_template}
        \vspace{-3pt}
\resizebox{0.95\textwidth}{!}{
\small

\begin{tabular}{lcc}
  \toprule
   & Best Value Found & Failure Prob. (\%) \\
  \midrule
  Original Prompt (Ours) & $0.056 \pm 0.002$ & 39.5 \\
  \midrule
  w/o HDBO Guidance & $0.068 \pm 0.007$ & 35.7 \\
  w/o Constraint Guidance & $0.064 \pm 0.003$ & 44.9 \\
  \bottomrule
  \end{tabular}

}
        
        
        \captionof{table}{Robustness on different LLMs. Experiments are conducted with four random seeds in the SVM benchmark.}
        \label{tab:seed_b}
\resizebox{0.85\textwidth}{!}{
\small
 \begin{tabular}{lcc}
  \toprule
   & Best Value Found & Failure Prob. (\%) \\
  \midrule
  GPT-4o (Ours) & 0.056 $\pm$ 0.002 & 39.5 \\
  \midrule
  GPT-4o-mini & 0.060 $\pm$ 0.003 & 43.1 \\
  GPT-5-mini & 0.056 $\pm$ 0.003 & 30.4 \\
  \bottomrule
  \end{tabular}

}

    \end{minipage}
    \vspace{-13pt}
\end{figure}

\vspace{3pt}
\textbf{Ablation on Each Component.} 
We conduct a systematic ablation study by removing each component. For the initial population, we replace BOCK and SL with Linear and Periodic kernels, matching the base kernel set used in CAKE. For the evaluation metric, we replace LOO-CRPS with the standard marginal log-likelihood (MLL). As shown in \Cref{fig:system_arch}, performance consistently drops as each component is removed. In particular, the initial population significantly affects sample efficiency, validating that kernels specifically designed for high-dimensional BO encode inductive biases that are valuable for effective surrogate modeling. We extend this ablation with more variants in \Cref{app:extend_ablation}.



\begin{figure}[t]
    \centering
    \begin{subfigure}[b]{0.48\textwidth}
        \centering
        \includegraphics[width=1\linewidth]{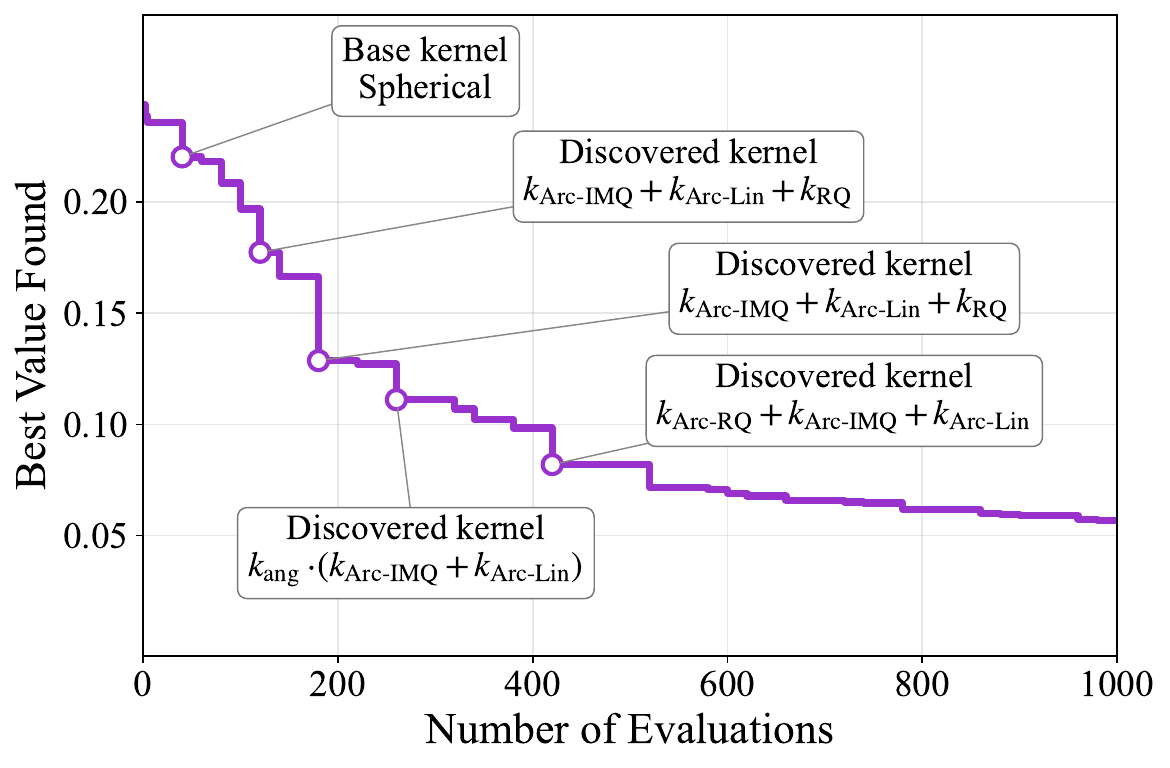}
        \caption{Summary of selected kernels for the top-5 performance improvements.}
        \label{fig:insights}
    \end{subfigure}
    \hfill 
    \begin{subfigure}[b]{0.48\textwidth}
        \centering
        \includegraphics[width=1\linewidth]{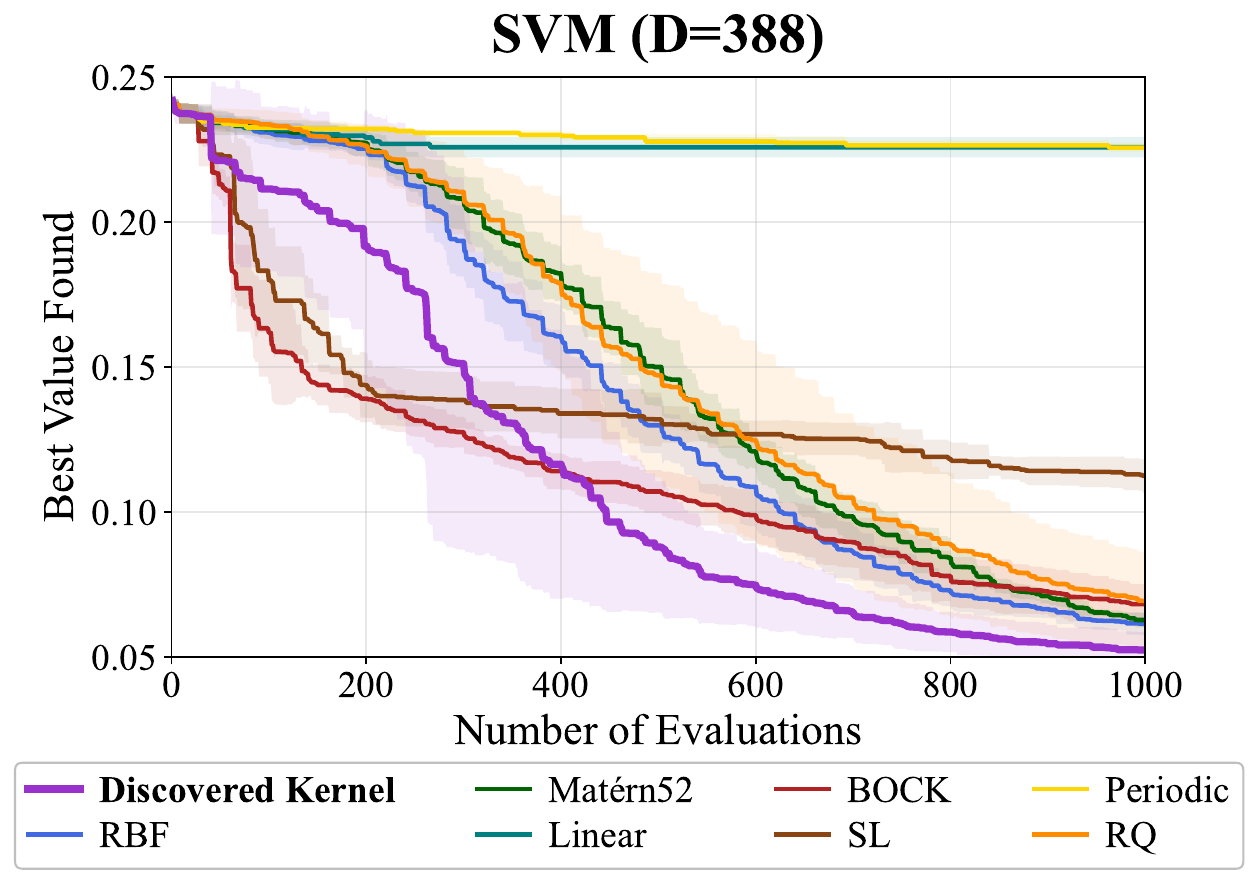}
        \caption{Evaluation of discovered kernel from SVM benchmark directly without kernel search.}
        \label{fig:transferability}
    \end{subfigure}
    \caption{Post-analysis on kernel discovery. Both experiments are conducted in SVM benchmark.}
    \vspace{-13pt}
\end{figure}

\vspace{3pt}
\textbf{Analysis on Prompt Template.}
As shown in \Cref{fig:KD_prompt}, our prompt includes two key guidelines for the LLM: HDBO and Constraint guidelines. We ablate each by removing it in turn. As shown in \Cref{tab:prompt_template}, removing the HDBO guideline degrades performance, whereas removing the Constraint guideline leads to a high failure rate.

\vspace{3pt}
\textbf{Robustness to Different LLMs.}
By default, we use \texttt{GPT-4o} for kernel formulation and code generation. To assess robustness across different LLMs, we run experiments with different LLMs: \texttt{GPT-4o-mini} and \texttt{GPT-5-mini}. As shown in \Cref{tab:seed_b}, we find no significant performance gap across models, though \texttt{GPT-4o-mini} has a slightly higher kernel rejection rate due to weaker code reasoning, while \texttt{GPT-5-mini} generates fewer invalid kernels. 
We also conduct ablation studies with open-source LLMs and different LLM backbones for other LLM-based methods in \Cref{app:robustness_llm}.

\section{Analysis of Discovered Kernels}
In this section, we characterize which kernels our pipeline discovers, how they evolve over training iterations, and whether they transfer across benchmarks.

\vspace{3pt}
\textbf{Evolution of Discovered Kernels.}
A key question is whether our pipeline discovers genuinely novel structures or merely recovers known base kernels. To investigate this, we conduct a systematic study of the kernels selected by our pipeline over training iterations in \Cref{fig:insights}. In the initial rounds, the pipeline favors simple base kernels with geometric input warpings (e.g., Spherical). As optimization progresses, more advanced structures emerge: the pipeline introduces geometric transformations such as \texttt{arctan} warping and composes them with existing kernels (e.g., RQ $\times$ Linear with \texttt{arctan}, further combined with BOCK or IMQ). This suggests that compositions of multiple geometrically-warped kernels are a promising direction for high-dimensional BO, going beyond the single-warping strategies explored by prior works. We provide detailed explanations of the discovered kernels, along with the same evolution analysis across all benchmarks, in \Cref{app:insights}.

\vspace{3pt}
\textbf{Transferability of Discovered Kernels}.
While our method performs kernel discovery for a single black-box function, we can extract kernels discovered by one benchmark and determine whether they can be transferred to other benchmarks. To verify this, we choose one kernel discovered from the SVM benchmark as follows:
\begin{equation}\label{eq:discover}
    k_{\text{discover}}(\mathbf{x}, \mathbf{x}') = k_{\text{Matérn52}}(\mathbf{x},\mathbf{x}')
\cdot
\left[
k_{\mathrm{tanh\text{-}Poly}}(\mathbf{x},\mathbf{x}')
+
k_{\mathrm{RQ}}(\mathbf{x},\mathbf{x}')
\right]
\end{equation}
Note that this kernel cannot be obtained via naive compositions since $k_{\text{poly}}$ does not exist in the base population and the $\texttt{tanh}(\cdot)$ operation is not allowed. As shown in \Cref{fig:transferability}, the discovered kernel outperforms base kernels on the SVM benchmark, and achieves an average rank of 2.8 across 8 fixed-kernel methods (full results in \Cref{tab:transferability_extended}).
To understand why the discovered kernel achieves superior performance on the SVM benchmark, we analyze the boundary hit ratio and the observation traveling salesman distance (OTSD) and find that the discovered kernel does not exhibit boundary-seeking behavior (See \Cref{fig:boundary_seeking}).
It is noteworthy that $k_{\text{tanh\text{-}Poly}}$ is a non-stationary kernel that is not typically preferred for BO, yet it performs well in high dimensions. This suggests that the role of non-stationary kernels in high-dimensional BO warrants further investigation. We provide detailed explanations of the discovered kernel and also present other frequently discovered kernels in \Cref{app:transferability}.


\section{Conclusion}

We presented Kernel Discovery, an LLM-driven evolutionary framework for designing effective GP kernels in high-dimensional BO. We adopt a two-stage pipeline in which one LLM proposes novel mathematical kernel forms and the other LLM converts them into executable code. To select among discovered kernels without overfitting to current observations, we introduce LOO-CRPS, an overfit-resistant criterion. On five high-dimensional BO benchmarks, our method achieves an average rank of 1.2 out of 17 methods. Beyond performance, our analysis of the discovered kernels provides new insights into what drives effective surrogate modeling in high-dimensional BO.

\vspace{3pt}
\textbf{Limitations and Future Work.}
Our current framework runs the discovery process independently for each benchmark, optimizing kernels for each benchmark. While we demonstrate that kernels discovered on one benchmark can transfer to others, a natural extension is to develop a general discovery strategy that operates across multiple benchmarks simultaneously, leveraging shared structure to amortize the search cost. Additionally, some kernels generated by LLMs do not pass the verifiers for positive semi-definiteness or dimension-agnosticity. We believe that the failure rate could be further reduced by detailed prompt engineering or by imposing constraints during generation, although we view these as engineering directions orthogonal to our main contribution, which is to demonstrate that LLM-driven kernel discovery is a promising strategy for high-dimensional BO.

\clearpage
\bibliography{neurips_2026.bib}
\bibliographystyle{unsrtnat}


\clearpage
\appendix
\section*{Appendix}

\section{Task Details}
\label{app:task_details}

We evaluate our method on five standard high-dimensional BO benchmarks covering a range of problem types and dimensionalities.

\paragraph{Rover ($D=100$).}
The Rover benchmark \citep{wang2018batched} is a trajectory optimization task in which a rover must navigate from a fixed start position to a fixed goal on a 2D terrain. The trajectory is parameterized by 50 intermediate 2D waypoints, yielding a $D=100$-dimensional continuous search space in $[0,1]^D$. The objective rewards proximity to the goal while penalizing path irregularity; higher values are better. The reward landscape contains many local optima corresponding to different trajectories.

\paragraph{Mopta08 ($D=124$).}
The Mopta08 benchmark is a structural car body engineering design problem \citep{eriksson2021high}. The task is to minimize the mass of a car structural component while satisfying 68 physical safety and manufacturing constraints. The 124-dimensional search space encodes continuous geometric design parameters. Constraint violations are aggregated into a penalty term, making the effective objective landscape highly non-smooth. Lower values are better.

\paragraph{Lasso-DNA ($D=180$).}
Lasso-DNA is from the LassoBench suite \citep{vsehic2022lassobench}, a benchmark collection for high-dimensional hyperparameter optimization of regularized regression models on genomic data. The input is a $D=180$-dimensional vector of group-wise LASSO regularization coefficients; the objective is the mean squared prediction error on a validation set from a DNA microarray gene expression study (lower is better). The structured sparsity of the optimal regularizer makes this benchmark representative of practical bioinformatics hyperparameter search.

\paragraph{SVM ($D=388$).}
The SVM benchmark \citep{eriksson2021high} involves hyperparameter optimization of a Support Vector Machine classifier on a high-dimensional tabular dataset. The $D=388$-dimensional search space encodes regularization and kernel hyperparameters of the SVM. The objective is the test misclassification error (lower is better). The high sensitivity of SVM performance to specific hyperparameter regions makes this a particularly challenging task.

\paragraph{Humanoid ($D=6392$).}
The Humanoid benchmark \citep{wang2020learning} is a continuous locomotion control task from the MuJoCo physics simulator. A simulated humanoid robot is controlled by a linear policy that maps the 376-dimensional state observation vector to 17 joint torques, yielding a weight matrix of size $17 \times 376 = D = 6392$ parameters. The objective is the cumulative reward accumulated over a fixed-length simulation episode (higher is better).

\paragraph{Ant ($D=888$, excluded from main evaluation).}
We first considered the Ant locomotion benchmark from MuJoCo, in which a quadruped robot is controlled by a linear policy mapping its state observations to joint torques. However, we excluded Ant from the main evaluation due to its abnormal reward structure. Specifically, we found out that the trivial zero-initialization baseline, which sets all policy parameters to $\mathbf{x} = \mathbf{0}$, achieves a cumulative reward of $+1000.40 \pm 4.40$. By contrast, all kernel-based GP-BO methods obtain dramatically lower or negative rewards, as shown in \Cref{tab:kernel_vs_baseline}. This results makes the benchmark unsuitable for comparing kernel quality in BO.

\begin{table}[h]
  \centering
  \caption{Final best values across base kernels compared to the zero-init-parameter policy baseline. Experiments are conducted with 4 random seeds.}
  \vspace{3pt}
  \label{tab:kernel_vs_baseline}
  \begin{tabular}{lc}
  \toprule
  Arm & Best Value Found \\
  \midrule
  $\mathbf{x}=\mathbf{0}$ (zero-init-parameter policy) & $\phantom{-}1000.40 \pm 4.40\phantom{00}$ \\
  \midrule
  RBF         & $-\phantom{0}778.67 \pm 176.89$ \\
  Matérn52    & $-\phantom{0}810.20 \pm 112.79$ \\
  RQ          & $-1011.04 \pm 261.02$ \\
  SL   & $-\phantom{0}660.03 \pm 86.11\phantom{0}$ \\
  BOCK & $\phantom{-}\phantom{0}151.70 \pm 358.00$ \\
  \bottomrule
  \end{tabular}
  \end{table}

\clearpage
\section{Baseline Details}
\label{app:baseline_details}

We compare against 16 baselines across three categories: base kernels, search-based methods, and LLM-based methods.

\subsection*{Base Kernels}

The following methods use the standard GP-BO setup with a single fixed kernel. For RBF, Matérn52, and RQ, we apply the data-dependent lengthscale prior of \citet{hvarfner2024vanilla}. Let $\mathbf{L} = \mathrm{diag}(\ell_1,\ldots,\ell_D)$ denote the ARD lengthscale matrix and $r_{\mathbf{L}} = \sqrt{(\mathbf{x}-\mathbf{x}')^\top\mathbf{L}^{-1}(\mathbf{x}-\mathbf{x}')}$. 

\begin{itemize}[leftmargin=2em]
    \item \textbf{RBF.} The squared exponential kernel with ARD:
    \begin{equation}
        k_{\mathrm{RBF}}(\mathbf{x},\mathbf{x}') = \exp\!\left(-\tfrac{1}{2}\,r_{\mathbf{L}}^2\right).
    \end{equation}

    \item \textbf{Matérn52.} The Matérn kernel with smoothness $\nu=5/2$ and ARD:
    \begin{equation}
        k_{\text{Matérn52}}(\mathbf{x},\mathbf{x}') = \left(1 + \sqrt{5}\,r_{\mathbf{L}} + \tfrac{5}{3}\,r_{\mathbf{L}}^2\right)\exp\!\left(-\sqrt{5}\,r_{\mathbf{L}}\right).
    \end{equation}
    Models twice mean-square differentiable functions; often better suited than RBF to practical optimization landscapes.

    \item \textbf{Linear.} The dot-product kernel with an unconstrained scale parameter $v \in \mathbb{R}$:
    \begin{equation}
        k_{\mathrm{Lin}}(\mathbf{x},\mathbf{x}') = v\,\mathbf{x}^\top\mathbf{x}'.
    \end{equation}

    \item \textbf{Periodic.} The ARD periodic kernel with per-dimension lengthscales $\ell_i > 0$ and periods $p_i > 0$:
    \begin{equation}
        k_{\mathrm{Per}}(\mathbf{x},\mathbf{x}') = \exp\!\left(-2\sum_{i=1}^{D}\frac{\sin^2\!\left(\pi(x_i - x_i')/p_i\right)}{\ell_i^2}\right).
    \end{equation}

    \item \textbf{RQ (Rational Quadratic).} A scale mixture of RBF kernels with Gamma-distributed inverse lengthscales:
    \begin{equation}
        k_{\mathrm{RQ}}(\mathbf{x},\mathbf{x}') = \left(1 + \frac{r_{\mathbf{L}}^2}{2\alpha}\right)^{-\alpha}, \quad \alpha > 0.
    \end{equation}
    The parameter $\alpha$ controls the relative contribution of short-range vs.\ long-range variation.

    \item \textbf{BOCK.} Bayesian Optimization with Cylindrical Kernels (BOCK) \citep{oh2018bock} decomposes each normalized input $\bar{\mathbf{x}} = (\mathbf{x}-\mathbf{c})/R$ into a radial component $r=\|\bar{\mathbf{x}}\|$ and an angular component $\hat{\mathbf{x}}=\bar{\mathbf{x}}/r$. The radial coordinate is transformed by the Kumaraswamy CDF $\kappa(r;\alpha,\beta)=1-(1-r^{\alpha})^{\beta}$ before applying a Matérn52 kernel, and the angular component is modeled by a polynomial kernel in the inner product:
    \begin{equation}
        k_{\mathrm{BOCK}}(\mathbf{x},\mathbf{x}') = k_{\text{Matérn52}}\!\left(\kappa(r),\kappa(r')\right)\cdot\sum_{p=0}^{P-1}w_p\,(\hat{\mathbf{x}}^\top\hat{\mathbf{x}}')^p,\quad w_p>0.
    \end{equation}
    This product structure separates the smooth radial variation from the angular geometry, mitigating boundary-seeking behavior.

    \item \textbf{SL (Spherical Linear).} The kernel of \citet{doumontwe} first applies ARD lengthscale scaling followed by a global scale $g$, then maps the result to the unit sphere via the stereographic projection $\psi:\mathbb{R}^D\to S^D$:
    \begin{equation}
        \psi(\mathbf{z}) = \frac{1}{1+\|\mathbf{z}\|^2}\begin{pmatrix}2\mathbf{z}\\\|\mathbf{z}\|^2-1\end{pmatrix}\in S^D.
    \end{equation}
    Letting $\tilde{\mathbf{x}} = (\mathbf{x}-\mathbf{c})\,/\,(\boldsymbol{\ell}\cdot g)$, the kernel is a softmax-weighted mixture of a linear term on the sphere and a constant bias:
    \begin{equation}
        k_{\mathrm{SL}}(\mathbf{x},\mathbf{x}') = \lambda_1\,\psi(\tilde{\mathbf{x}})^\top\psi(\tilde{\mathbf{x}}') + \lambda_0, \quad \lambda_0+\lambda_1=1,\;\lambda_0,\lambda_1\geq 0.
    \end{equation}
    The stereographic projection ensures that inputs at any distance from the origin are mapped to a bounded region on the sphere, avoiding the unbounded extrapolation that afflicts linear kernels in high dimensions.
\end{itemize}

\subsection*{Search-based Methods}
All search-based baselines share the same initial population $\mathcal{P}_0$ as our method for a fair comparison.

\begin{itemize}[leftmargin=2em]
    \item \textbf{Greedy Search.} At each BO iteration, it evaluates every kernel in $\mathcal{P}_0$ and selects the one with the lowest Bayesian Information Criterion (BIC). No new kernels are created.

    \item \textbf{Compositional Search.} At each BO iteration, use greedy search to discover kernel structures that best explain the observed data via pairwise additive and multiplicative compositions of kernels in the current population. Following the paper, we use BIC as a selection metric and directly use hyperparameters fitted in a single kernel to composite kernels without refitting. We use the max depth of the composition as $3$.

    \item \textbf{CAKE.} Context-Aware Kernel Evolution \citep{suwandiadaptive} uses an LLM to select and compose base kernels conditioned on prior observations. Its search is restricted to additive and multiplicative compositions of a fixed base set, and passing raw observations to the LLM induces context-length issues in high dimensions. If the dimensions are too high, we truncate the observations to prevent context limit violation. We use the max depth of the composition as $3$. While CAKE uses \texttt{GPT-4o-mini} for the LLM API, we evaluate the baseline with \texttt{GPT-4o} for a fair comparison.
\end{itemize}

\subsection*{LLM-based Methods}
For all LLM-based baselines, we initialize the dataset with Sobol sequences of size of $20$ as our method for a fair comparison.

\begin{itemize}[leftmargin=2em]
    \item \textbf{LLaMEA-BO (ARM, ATRBO, TREvol, TROpt, TRPareto).} Five end-to-end LLM-driven BO algorithms produced by the LLaMEA-BO pipeline \citep{li2025llamea}, which uses an LLM in an evolutionary loop to generate and refine BO algorithm code. Each variant is a distinct algorithm discovered by the pipeline. These methods may not use a GP surrogate and directly propose candidates from prior observations.

    \item \textbf{LMABO.} An LLM-based BO method that adaptively selects the acquisition function at each iteration \citep{ngoadaptive}. For a fair comparison, we use GP with an RBF as the base kernel, following \citet{hvarfner2024vanilla}, and use \texttt{L-BFGS-B} as the optimizer.

    \item \textbf{LLAMBO (excluded from main evaluation).}
    LLAMBO \citep{liularge} uses LLMs to propose and evaluate candidates conditioned on historical evaluations, making its prompt size scale with both history length and input dimensionality. On Rover (\(D=100\)), the full history of 980 evaluations already requires approximately 600K input tokens, exceeding the 128k context window of GPT-4o. Even after truncating the history to 20 observations, our GPT-4o probe costs about \(\$1.05\) per BO iteration, corresponding to about \(\$1{,}024\) for a single full run. We therefore exclude LLAMBO due to context-length and budget limitations.
\end{itemize}

\paragraph{Additional Baselines: TuRBO and SAASBO.}
\label{app:additional_baselines}
We compare against two additional baselines, TuRBO \citep{eriksson2019scalable} and SAASBO \citep{eriksson2021high}, which are not included in the main table as they employ different acquisition function maximization strategies and inference procedures (TuRBO uses a trust-region approach and SAASBO uses sparse axis-aligned priors with MCMC-based inference). However, we compare the results of those baselines as they are widely used in high-dimensional BO. As shown in \Cref{tab:additional_baselines}, Kernel Discovery outperforms both methods across all benchmarks.

\begin{table*}[h]
\centering
\caption{Comparison with additional baselines across standard benchmarks. SAASBO results use $T=500$ due to its higher computational cost. Experiments are conducted with 4 random seeds.}
\label{tab:additional_baselines}
\vspace{2pt}
\renewcommand{\arraystretch}{1.2}
\resizebox{\textwidth}{!}{
\begin{tabular}{l ccccc}
\toprule
\multirow{2}{*}{\textbf{Method}} & \textbf{Rover $(\uparrow)$} & \textbf{Mopta08 $(\downarrow)$} & \textbf{Lasso-DNA $(\downarrow)$} & \textbf{SVM $(\downarrow)$} & \textbf{Humanoid $(\uparrow)$} \\
& ($D=100$) & ($D=124$) & ($D=180$) & ($D=388$) & ($D=6392$) \\
\midrule
\textbf{Kernel Discovery (Ours)} & \textbf{\textcolor{black}{4.353 $\pm$ 0.319}} & \textbf{\textcolor{black}{216.81 $\pm$ 0.87}} & \textbf{\textcolor{black}{0.286 $\pm$ 0.001}} & \textbf{\textcolor{black}{0.056 $\pm$ 0.003}} & \textbf{\textcolor{black}{762.78 $\pm$ 72.39}} \\
\midrule
TuRBO & 3.584 $\pm$ 0.175 & 242.39 $\pm$ 3.60 & 0.289 $\pm$ 0.001 & 0.142 $\pm$ 0.015 & 449.26 $\pm$ 85.56 \\
SAASBO ($T=500$) & 3.981 $\pm$ 0.362 & 223.30 $\pm$ 3.44 & 0.290 $\pm$ 0.001 & 0.148 $\pm$ 0.012 & 449.08 ± 53.58 \\
\bottomrule
\end{tabular}
}
\end{table*}

\clearpage
\section{Implementation Details}
\label{app:implement_detail}

\paragraph{Shared BO Setup.}
For all methods except end-to-end LLM-based approaches (ARM, ATRBO, TREvol, TROpt, TRPareto), we use a shared GP-based BO framework built on GPyTorch and BoTorch. The surrogate is an exact Gaussian Process with homoscedastic Gaussian observation noise. Kernel hyperparameters are optimized by maximizing the marginal log-likelihood (MLL) using \texttt{L-BFGS-B} with 3 random restarts. Inputs are normalized to $[0,1]^D$ and outputs are standardized to zero mean and unit variance before fitting.

We use \texttt{LogEI} as the acquisition function and optimize it using \texttt{L-BFGS-B} with 4 random restarts initialized from 512 Sobol quasi-random candidates. RAASP sampling is implemented by tuning the option \texttt{sample\_around\_best=True}. The initial dataset $\mathcal{D}_0$ is generated by a Sobol sequence of size $|\mathcal{D}_0| = 20$. The total evaluation budget is $T = 1000$ function queries.

All experiments use a batch BO setting with batch size $q = 20$. At each BO round, the acquisition function is maximized to select a batch of $q$ candidate points, which are evaluated simultaneously. This yields $T/q = 50$ BO rounds per run. All baselines also use the same $q = 20$ batch size to ensure a fair comparison. Please refer to \Cref{tab:batch_size_comparison} for the comparison of the performance of base kernels under different batch sizes. All experiments are done with a single RTX NVIDIA L40S GPU.

\lstdefinestyle{promptstyle}{
    backgroundcolor=\color{black!5},
    frame=single,
    breaklines=true,                   
    breakautoindent=false,             
    breakindent=0pt,                   
    captionpos=b,
    keepspaces=true,
    numbers=left,
    numberstyle=\tiny\color{gray},
    basicstyle=\ttfamily\footnotesize,
    breaklines=true,
}

\paragraph{Full Prompt Templates.}
We present the full prompt templates for each stage, mathematical form discovery, code conversion, and composition, in \Cref{fig:prompt_dis,fig:prompt_conv,fig:prompt_comp}, respectively. In addition to generating solely mathematical expressions or code blocks, we instruct LLM to justify its answer for exploring its reasoning capabilities and the robustness of the pipeline.

\begin{figure}[h!]

\begin{nolinenumbers}
\begin{lstlisting}[style=promptstyle, caption={Mathematical form discovery prompt template}, label={fig:prompt_dis}]
You are an expert in Gaussian process kernel design for high-dimensional Bayesian optimization.

TOP {len(parent_formulas_with_loss)} kernels in our population:
{context}

Generate {n_discoveries} DISTINCT, CREATIVE, mathematically innovative DISCOVERED kernels.
Each discovery should be derived from one or more of the above kernels, but with meaningful mathematical changes.
{highdim}{psd}{simplicity}
OUTPUT FORMAT:
Return EXACTLY {n_discoveries} kernel descriptions. Enclose EACH in a ```formula block:

```formula
KERNEL: [name]

PARAMETERS:
- [name] ([shape], [constraint]): [description]

INPUT TRANSFORM:
  [step-by-step math]

COVARIANCE FUNCTION:
  [k(x1, x2) = ...]

PSD GUARANTEE:
  [why this kernel is PSD]
```

Each MUST be fundamentally distinct (different transform, different covariance structure, different mathematical idea)."""
\end{lstlisting}
\end{nolinenumbers}
\end{figure}

\clearpage
\begin{nolinenumbers}
\begin{lstlisting}[style=promptstyle, caption={Code conversion prompt template.}, label={fig:prompt_conv}]
You are a GPyTorch kernel code generator. Convert this kernel formula into a GPyTorch EvolvedKernel class.

{formula}

REFERENCE TEMPLATE A - Distance-based kernel {RBF code}

REFERENCE TEMPLATE B - Feature-map kernel {SL code}

CRITICAL RULES:
1. __init__ SIGNATURE: `__init__` MUST accept `ard_num_dims: int` as the ONLY required parameter. The system calls `EvolvedKernel(ard_num_dims=D)`. All other args MUST have defaults. NEVER add required args like `q: int`, `center: Tensor`, `M: int`.
2. FORWARD SIGNATURE: `def forward(self, x1, x2, diag=False, **params)` - always accept **params.
3. BATCH-SAFE SHAPES (CRITICAL): The kernel is tested with these exact shapes - your code MUST handle ALL of them:
   - 2D: x1=(5, D) vs x2=(1, D) -> output must be (5, 1), NOT (5, 5)
   - 2D: x1=(3, D) vs x2=(7, D) -> output must be (3, 7)
   - 3D batched (BoTorch acqf optimization): x1=(1, 4, D) vs x2=(1, 3, D) -> output must be (1, 4, 3)
   Therefore: ALL dim indexing must be relative (-1 for D, -2 for N). NEVER use `.size(0)`, `.expand(x.size(0), ...)`, or `.shape[0]`. For broadcasting scalars to match batch+N dims, use `torch.ones_like(x[..., :1])`.
4. OUTPUT SHAPE: return (..., N1, N2) for ANY N1 != N2. If your output is (N1, N1) instead of (N1, N2), you have a bug.
5. DISTANCE: `torch.cdist(x1s, x2s, p=2)` - NEVER `x1.unsqueeze(-2) - x2.unsqueeze(-3)` (OOM).
6. INNER PRODUCT: `x1 @ x2.transpose(-1, -2)`.
7. DEVICE: new tensors in forward use `device=x1.device`. Use `register_buffer` in __init__.
8. NO IN-PLACE: `x = x + y` not `x += y`. `x = x.clamp(...)` not `x.clamp_(...)`.
9. NO DETACH/NUMPY: never `.detach()`, `.numpy()`, `.item()`, `torch.no_grad()` in forward.
10. NUMERICAL: `.clamp(min=1e-15)` before sqrt/log, `.clamp(max=20.0)` for exp.
11. LENGTHSCALE: if `has_lengthscale=True`, do NOT `register_parameter("raw_lengthscale", ...)`.
12. PARAM ORDER: `register_parameter(name, ...)` BEFORE `register_constraint(name, ...)`.
13. DIAG: `if diag: return covar.diagonal(dim1=-2, dim2=-1)`.
14. IMPORTS: include `import math` if using math.sqrt/pi/log.
15. NO SELF-REASSIGNMENT in forward(): never `self.x = ...` - use local variables.
16. ONLY USE EXISTING gpytorch classes.
17. NO SUB-KERNELS: do NOT instantiate gpytorch kernel objects (e.g. `MaternKernel()`, `RBFKernel()`) as sub-components. They produce LazyEvaluatedKernelTensor with unpredictable shapes. Instead, implement the formula directly (e.g. for Matern-5/2: `(1 + sqrt5*d + 5/3*d**2) * exp(-sqrt5*d)`).
18. NO DATA-DEPENDENT BOUNDS in forward(): never compute bounds/center from the input data (e.g. `x1.max(dim=-2)` or `x1.min(dim=-2)`). These change with N and break shape invariance. Store fixed bounds via `register_buffer` in __init__, or use constants (e.g. 0 and 1).
19. SCALAR PAIRWISE DISTANCE: when computing distance between scalar values (e.g. warped radii of shape (..., N, 1)), use `torch.cdist(r1, r2, p=2)` which gives (..., N1, N2). NEVER use `r1 - r2` or `torch.abs(r1 - r2)` - this gives (..., N1, 1) not (..., N1, N2) and breaks cross-pair computation.

Return ONLY the Python code in a ```python block."""
\end{lstlisting}
\end{nolinenumbers}
\clearpage
\begin{nolinenumbers}
\begin{lstlisting}[style=promptstyle, caption={Composition prompt template.}, label={fig:prompt_comp}]
You are combining GP kernels for high-dimensional Bayesian optimization.

Population of {len(population_formulas_with_loss)} kernels:
{context}

Generate {n_compositions} DISTINCT composed kernels. For EACH composition:
- Pick exactly 2 kernels from the population that COMPLEMENT each other mathematically
- Combine them: sum their covariance functions, multiply them, share a transform with different distance metrics, or any other valid composition
- Explain WHY these two kernels complement each other
{psd}{simplicity}
OUTPUT FORMAT:
Return EXACTLY {n_compositions} kernel descriptions. Enclose EACH in a ```formula block:

```formula
KERNEL: [name]
COMPOSED FROM: [Kernel X] + [Kernel Y]  (or *, or other operation)
WHY: [1 sentence on why these complement each other]

PARAMETERS:
- [param] ([shape], [constraint]): [description]

INPUT TRANSFORM:
  [step-by-step math]

COVARIANCE FUNCTION:
  k(x1, x2) = ...

PSD GUARANTEE:
  [why this kernel is PSD]
```"""

\end{lstlisting}
\end{nolinenumbers}
\vspace{-12pt}


\paragraph{Kernel Validation Failure Rates.}
\label{app:failure_rates}
The failure probability in \Cref{tab:prompt_template,tab:seed_b} denotes the empirical fraction of LLM-generated kernels rejected by our structural validator. A generated kernel \(k_{\mathrm{new}}\) is accepted only when
\[
\mathcal{V}(k_{\mathrm{new}})
=
\mathcal{V}_{\mathrm{agn}}(k_{\mathrm{new}})
\wedge
\mathcal{V}_{\mathrm{psd}}(k_{\mathrm{new}})
= 1.
\]
Here, \(\mathcal{V}_{\mathrm{agn}}\) checks whether the generated kernel implementation is dimension- and shape-agnostic, as required by the BO pipeline. Concretely, we instantiate the kernel with different input dimensionalities and evaluate its \texttt{forward} function on several synthetic input configurations, including self-covariance inputs \((N,D)\), cross-covariance inputs \((N_1,D)\) and \((N_2,D)\) with \(N_1 \neq N_2\), and batched inputs used during acquisition optimization. The kernel is rejected if any call raises a runtime error, assumes a hard-coded dimensionality, fails to broadcast correctly, or returns a covariance tensor with an invalid shape. The PSD check \(\mathcal{V}_{\mathrm{psd}}\) then evaluates the kernel on random inputs and tests whether the resulting Gram matrix is numerically positive semi-definite via a Cholesky decomposition with a small jitter. We report the fraction of generated kernels for which at least one of these checks fails, aggregated over BO iterations and random seeds.

\paragraph{Effect of Batch Size ($q$).}
The main experiments use $q=20$, reducing the number of BO rounds to $T/q = 50$. To assess whether this setting disproportionately favors our method, we compare base kernels under $q=20$ vs.\ $q=1$ on the SVM benchmark. \Cref{tab:batch_size_comparison} shows that the relative ordering of kernels is largely preserved across both settings, and that the optimal kernel (RBF in the $q=1$ setting) is the same as in the $q=20$ setting. This confirms that the batch setting does not systematically alter which kernels are competitive, and that our comparison is not confounded by the choice of $q$.
\vspace{-35pt}
\begin{table}[h]
  \centering
  \caption{Comparison of final best values across base kernels under different oracle query batch sizes.}
  \label{tab:batch_size_comparison}
  \resizebox{\linewidth}{!}{%
  \begin{tabular}{lccccccc}
  \toprule
  & RBF & Matérn52 & RQ & BOCK & SL & Linear & Periodic \\
  \midrule
  $q=20$ & 0.061 $\pm$ 0.003 & 0.062 $\pm$ 0.002 & 0.069 $\pm$ 0.014 & 0.068 $\pm$ 0.006 & 0.111 $\pm$ 0.006 & 0.226 $\pm$ 0.003 & 0.226 $\pm$ 0.001 \\
  $q=1$  & 0.078 $\pm$ 0.003 & 0.117 $\pm$ 0.021 & 0.078 $\pm$ 0.007 & 0.070 $\pm$ 0.009 & 0.064 $\pm$ 0.002 & 0.226 $\pm$ 0.002 & 0.223 $\pm$ 0.002 \\
  \bottomrule
  \end{tabular}}
\end{table}

\clearpage
\section{Algorithm Details}
\label{app:algorithm}

\paragraph{Discard Logic.} The difference between the standard evolutionary pipeline and our task is that one kernel that suits a certain round may not suit the other rounds, as the data keeps changing. To mitigate this, we discard the kernel if it is selected but does not improve the best value found so far. For kernels in the initial population, which are the basic building blocks of our discovery pipeline, we discard them if they are selected and no improvement is observed after $P=3$ iterations. If all kernels in the population are discarded, we reset the population to the initial population.

\paragraph{Humanoid Benchmark.} In the humanoid benchmark, the dimension is too high, and we observed that even with the LOO-CRPS evaluation metric, it sometimes leads to selecting too complex kernels that may not lead to an improvement. To remedy this, we regularize the LOO-CRPS score by the BIC metric as follows: 
\begin{equation}
    \begin{aligned}
        \text{LOO-CRPS-BIC}(\mathbf{K}_k, \mathbf{y}) 
        \;=\; \frac{1}{n} \sum_{i=1}^{n} \text{CRPS}\!\left(\mathcal{N}(\mu_{-i},\, \sigma^2_{-i}),\, y_i\right) 
        \;+\; \frac{|\boldsymbol{\theta}_k| \log n}{n},
    \end{aligned}
\end{equation}
where $|\boldsymbol{\theta}_k|$ is the number of hyperparameters in the kernel $k$. We found that this regularization works well, especially in extremely high dimensions, $D \gg n$. 
\begin{algorithm}[h]
\caption{Kernel Discovery for High-Dimensional BO}
\label{alg:main}
\begin{algorithmic}[1]
\REQUIRE Objective $f$, initial dataset $\mathcal{D}_0$, population size $N$, BO budget $T$, initial-population patience $P$, LLM $\mathcal{M}$ with prompts $\rho_{\text{disc}}, \rho_{\text{conv}}, \rho_{\text{comp}}$, verifier $\mathcal{V}$
\STATE Initialize $\mathcal{P}_0 \leftarrow \{k_{\text{RBF}},\, k_{\text{Mat52}},\, k_{\text{RQ}},\, k_{\text{BOCK}},\, k_{\text{SL}}\}$, \quad $\mathcal{P} \leftarrow \mathcal{P}_0$
\STATE $\mathcal{D} \leftarrow \mathcal{D}_0$, \quad $y^* \leftarrow \max_{(\mathbf{x},y)\in\mathcal{D}_0} y$, \quad $\text{fail}[k] \leftarrow 0$ for all $k \in \mathcal{P}_0$
\FOR{$t = 0, 1, \ldots, T - 1$}
    \STATE \textbf{// Discovery stage}
    \STATE $m^{\text{new}} \sim \mathcal{M}\!\left(\{m_i\}_{k_i \in \mathcal{P}};\, \rho_{\text{disc}}\right)$, \quad $c^{\text{new}} \leftarrow \mathcal{M}(m^{\text{new}};\, \rho_{\text{conv}})$, \quad $k^{\text{new}} \leftarrow (m^{\text{new}}, c^{\text{new}})$
    \IF{$\mathcal{V}(k^{\text{new}}) = 1$ \textbf{and} GP fitting time $\leq 60$s}
        \STATE $\mathcal{P} \leftarrow \mathcal{P} \cup \{k^{\text{new}}\}$
    \ENDIF
    \STATE \textbf{// Composition stage}
    \STATE $m^{\text{comp}} \sim \mathcal{M}\!\left(\{m_{(i)}\}_{i=1}^{N};\, \rho_{\text{comp}}\right)$, \quad$c^{\text{comp}} \leftarrow \mathcal{M}(m^{\text{comp}};\, \rho_{\text{conv}})$, \quad $k^{\text{comp}} \leftarrow (m^{\text{comp}}, c^{\text{comp}})$
    \IF{$\mathcal{V}(k^{\text{comp}}) = 1$ \textbf{and} GP fitting time $\leq 60$s}
        \STATE $\mathcal{P} \leftarrow \mathcal{P} \cup \{k^{\text{comp}}\}$
    \ENDIF
    \STATE \textbf{// Kernel selection and BO step}
    \STATE $k^\star \leftarrow \arg\min_{k \in \mathcal{P}}\, \text{LOO-CRPS}(\mathbf{K}_k, \mathbf{y})$
    \STATE Fit GP with $k^\star$ on $\mathcal{D}$; \quad $\mathbf{x}_{t+1} \leftarrow \arg\max_{\mathbf{x}}\, \alpha(\mathbf{x};\, \text{GP})$
    \STATE $y_{t+1} \leftarrow f(\mathbf{x}_{t+1})$; \quad $\mathcal{D} \leftarrow \mathcal{D} \cup \{(\mathbf{x}_{t+1},\, y_{t+1})\}$
    \STATE \textbf{// Discard logic}
    \IF{$y_{t+1} \leq y^*$}
        \STATE $\text{fail}[k^\star] \leftarrow \text{fail}[k^\star] + 1$
        \IF{$k^\star \notin \mathcal{P}_0$ \textbf{or} $\text{fail}[k^\star] \geq P$}
            \STATE $\mathcal{P} \leftarrow \mathcal{P} \setminus \{k^\star\}$
        \ENDIF
    \ENDIF
    \IF{$\mathcal{P} = \emptyset$}
        \STATE $\mathcal{P} \leftarrow \mathcal{P}_0$; \quad $\text{fail}[k] \leftarrow 0$ for all $k \in \mathcal{P}_0$
    \ENDIF
    \STATE \textbf{// Population update}
    \STATE $\mathcal{P} \leftarrow \operatorname{Top\text{-}N}\!\left(\mathcal{P};\, \text{LOO-CRPS}(\cdot,\, \mathbf{y})\right)$, $y^* \leftarrow \max(y^{*}, y_{t+1})$
\ENDFOR
\STATE \textbf{return} $\arg\max_{(\mathbf{x},y)\in\mathcal{D}}\, y$
\end{algorithmic}
\end{algorithm}

\clearpage
\section{Extended Experiment Results}
\label{app:extend_results}

\begin{figure}[h]
    \centering
    \includegraphics[width=\linewidth]{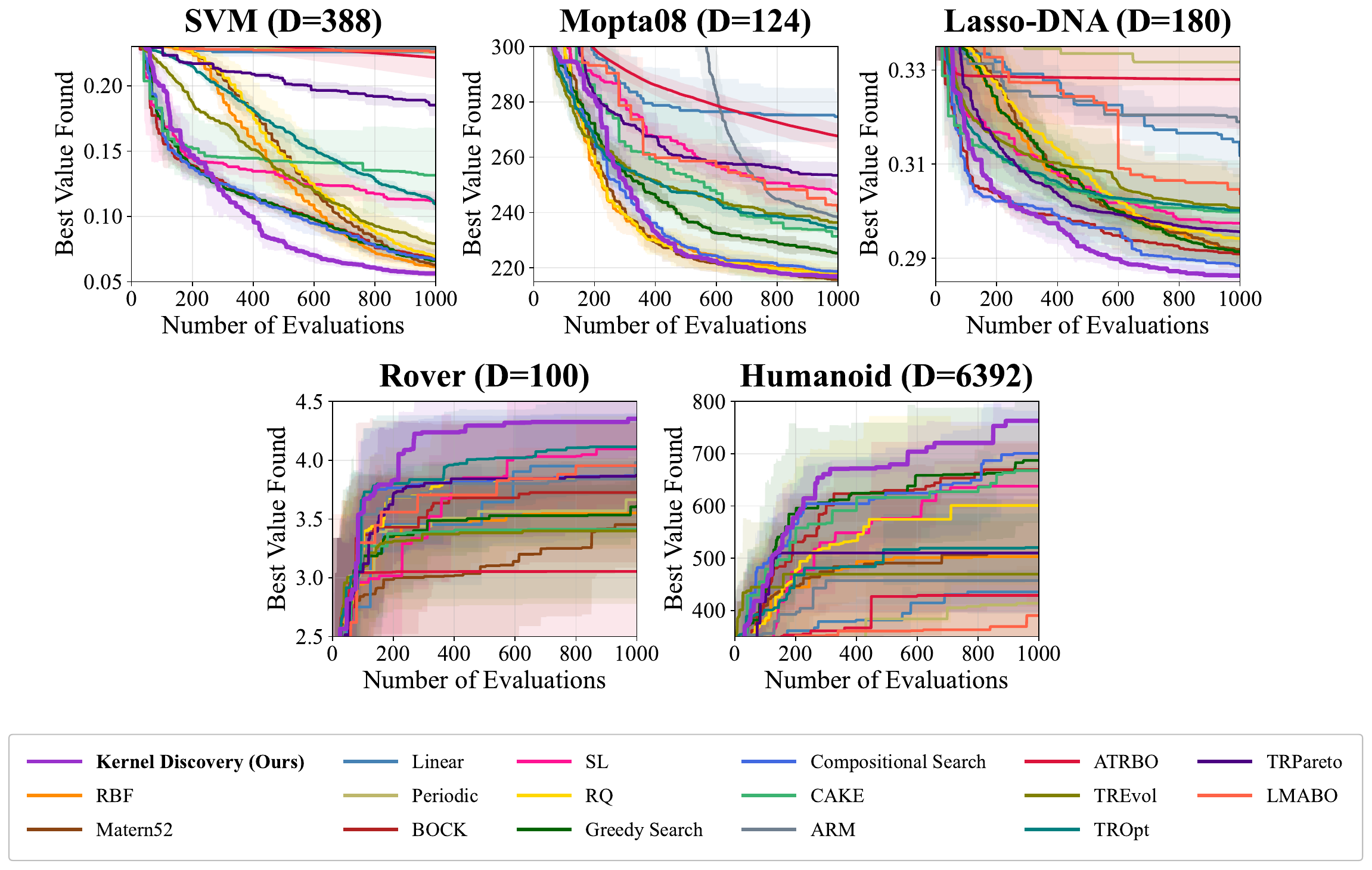}
    \caption{Full learning curves for all methods across five benchmarks.}
    \label{fig:main_extended}
\end{figure}

\paragraph{Full Learning Curves.}
\Cref{fig:main_extended} extends \Cref{fig:main} to show the full learning curves for all 16 baselines across all five benchmarks.

\clearpage
\section{Time Complexity Analysis}
\label{app:time_complexity}
We analyze the per-iteration wall-clock time of Kernel Discovery and compare it against all baselines in Table~\ref{tab:timing}. To systematically identify the bottleneck of our running time, we decompose each iteration into the following phases:
\begin{itemize}[leftmargin=1em]
    \item \textbf{GP Fitting}: Fitting all kernels in the current population in parallel using a process pool. The kernel selection score (e.g., LOO-CRPS) is computed within this phase, so the cost of choosing the best candidate is already included.
    \item \textbf{LLM Inference}: Two API calls per iteration - one for generating a novel mathematical kernel form and one for converting that form into executable code.
    \item \textbf{Kernel Validation}: Lightweight structural checks on each newly generated kernel before it is admitted to the population.
    \item \textbf{Search}: An additional kernel search procedure that applies only to search-based methods. Compositional Search enumerates pairwise additive and multiplicative compositions, while CAKE employs an LLM-guided composition with a BAKER-based selection procedure. Our method and Greedy Search select the best candidate directly from the evaluated pool, so this cost is negligible.
    \item \textbf{Acquisition Function Optimization}: Maximizing the acquisition function via \texttt{L-BFGS-B} to obtain the next query point.
    \item \textbf{Oracle Evaluation}: Querying the black-box benchmark function at the selected point, shared identically across all methods.
\end{itemize}
For LLM-based end-to-end methods, the pipeline does not follow the standard BO decomposition, so we report only the total per-iteration time and oracle evaluation cost.
As shown in Table~\ref{tab:timing}, our method incurs longer time than fixed base kernel methods but achieves comparable total running time to search-based methods. Among LLM-based baselines, TREvol and TRPareto require significantly more time per iteration. We also observe that LLM inference is the primary bottleneck in our pipeline, partially due to the two-stage approach that issues separate API calls for formula generation and code conversion. While this decomposition is critical for producing functionally diverse kernels (Section~6), developing more efficient inference strategies is a promising direction for future work.
\begin{table}[h]
\centering
\caption{Per-iteration wall-clock time breakdown (mean $\pm$ std, seconds) on the SVM benchmark with 4 random seeds. For LLM-based methods, individual phases are not separately tracked; we report only the total time and oracle evaluation time.}
\label{tab:timing}
\resizebox{\linewidth}{!}{
\begin{tabular}{lrrrrrrr}
\toprule
Method & GP Fit & LLM & Validation & Search & Acqf Opt & Oracle & Total \\
\midrule
\multicolumn{8}{l}{\textbf{Base Kernels}} \\
\midrule
RBF         & $0.68 \pm 0.02$ & N/A & N/A & $0.00 \pm 0.00$ & $2.70 \pm 0.97$ & $0.28 \pm 0.04$ & $5.18 \pm 1.00$  \\
Matérn52    & $0.69 \pm 0.04$ & N/A & N/A & $0.00 \pm 0.00$ & $2.76 \pm 0.99$ & $0.28 \pm 0.04$ & $5.26 \pm 1.00$  \\
Linear      & $0.95 \pm 0.58$ & N/A & N/A & $0.00 \pm 0.00$ & $3.14 \pm 0.88$ & $0.29 \pm 0.04$ & $6.49 \pm 2.38$  \\
Periodic    & $0.85 \pm 0.20$ & N/A & N/A & $0.00 \pm 0.00$ & $1.35 \pm 1.52$ & $0.25 \pm 0.04$ & $4.28 \pm 1.40$  \\
BOCK        & $1.14 \pm 0.44$ & N/A & N/A & $0.00 \pm 0.00$ & $4.03 \pm 1.62$ & $0.31 \pm 0.06$ & $7.02 \pm 1.87$  \\
SL          & $1.41 \pm 0.85$ & N/A & N/A & $0.00 \pm 0.00$ & $3.57 \pm 1.10$ & $0.30 \pm 0.04$ & $6.83 \pm 1.69$  \\
\midrule
\multicolumn{8}{l}{\textbf{Search-based}} \\
\midrule
Greedy Search        & $5.38 \pm 1.80$ & N/A             & N/A           & $0.00 \pm 0.00$   & $5.70 \pm 0.56$  & $0.35 \pm 0.05$ & $11.51 \pm 2.04$  \\
Compositional Search & $4.90 \pm 1.55$ & N/A             & N/A           & $13.80 \pm 6.78$  & $5.88 \pm 1.33$  & $0.34 \pm 0.05$ & $20.70 \pm 8.87$  \\
CAKE                 & $7.15 \pm 3.61$ & $9.09 \pm 3.23$ & N/A           & $17.44 \pm 3.73$  & $19.34 \pm 4.02$ & $0.35 \pm 0.06$ & $37.14 \pm 4.47$  \\
\midrule
\multicolumn{8}{l}{\textbf{LLM-based}} \\
\midrule
ARM      & N/A & N/A & N/A & N/A & N/A & $0.30 \pm 0.04$ & $4.54 \pm 0.02$    \\
ATRBO    & N/A & N/A & N/A & N/A & N/A & $0.28 \pm 0.05$ & $14.51 \pm 0.06$   \\
TREvol   & N/A & N/A & N/A & N/A & N/A & $0.31 \pm 0.02$ & $209.88 \pm 3.40$  \\
TROpt    & N/A & N/A & N/A & N/A & N/A & $0.29 \pm 0.05$ & $3.67 \pm 0.10$    \\
TRPareto & N/A & N/A & N/A & N/A & N/A & $0.30 \pm 0.02$ & $66.29 \pm 0.26$   \\
LMABO    & $3.56 \pm 0.50$ & $1.92 \pm 0.45$ & N/A & $0.00 \pm 0.00$ & $1.92 \pm 0.28$ & $0.27 \pm 0.04$ & $5.76 \pm 0.61$ \\
\midrule
\textbf{Kernel Discovery (Ours)} & $4.02 \pm 1.22$ & $14.24 \pm 2.05$ & $1.90 \pm 0.78$ & $0.00 \pm 0.00$ & $4.56 \pm 0.11$ & $0.30 \pm 0.01$ & $28.05 \pm 1.81$ \\
\bottomrule
\end{tabular}
}
\end{table}

\clearpage

\section{Direct Code Generation Analysis}
\label{app:direct_code_gen}

\paragraph{Details on Cosine Distance Metric.}
To quantify the functional diversity of generated kernels, we compare their induced Gram matrices on a shared reference input set. For each benchmark and each method, we collect all valid LLM-generated kernels produced during the BO run. We then sample a fixed reference batch
\(\mathbf{X}_{\mathrm{ref}} \in [0,1]^{N \times D}\) with \(N=80\), where \(D\) is the dimensionality of the corresponding benchmark. The same reference batch is used for all kernels within a benchmark.

For each generated kernel \(k_i\), we compute its Gram matrix
$K_i = k_i(\mathbf{X}_{\mathrm{ref}}, \mathbf{X}_{\mathrm{ref}}) \in \mathbb{R}^{N \times N},$
and flatten it into a vector
$\mathbf{v}_i = \mathrm{vec}(K_i) \in \mathbb{R}^{N^2}.$
We then compute the mean pairwise cosine similarity among all generated kernels in the pool \(\mathcal{P}\):
\[
\bar{s}
=
\frac{2}{|\mathcal{P}|(|\mathcal{P}|-1)}
\sum_{i<j}
\frac{\mathbf{v}_i^\top \mathbf{v}_j}
{\|\mathbf{v}_i\|_2 \|\mathbf{v}_j\|_2}.
\]
Finally, we report the cosine distance as $\mathrm{CosineDistance} = 1 - \bar{s}$. A higher value indicates that the generated kernels induce more diverse covariance structures on the same reference inputs.


\paragraph{Performance Comparison.}
We further compare direct code generation with our two-stage generation under the same experimental setup. As shown in \Cref{tab:cosine_distance_perf}, the two-stage approach achieves better final performance on all benchmarks. This suggests that generating kernels through an explicit mathematical formulation step leads to more functionally meaningful and consistently effective kernel candidates than directly generating code.
\begin{table*}[h]
\centering
\caption{Performance comparison between direct code generation and our two-stage generation.
Results are averaged over 4 random seeds.
Bold denotes the best entry in each column.}
\label{tab:cosine_distance_perf}
\vspace{2pt}
\renewcommand{\arraystretch}{1.2}

\resizebox{.95\textwidth}{!}{
\begin{tabular}{l ccccc}
\toprule
\multirow{2}{*}{\textbf{Method}} & \textbf{Rover $(\uparrow)$} & \textbf{Mopta08 $(\downarrow)$} & \textbf{Lasso-DNA $(\downarrow)$} & \textbf{SVM388 $(\downarrow)$} & \textbf{Humanoid $(\uparrow)$} \\
& ($D=100$) & ($D=124$) & ($D=180$) & ($D=388$) & ($D=6392$) \\
\midrule

\textbf{Two-Stage Generation (Ours)} & \textbf{4.353 $\pm$ 0.319} & \textbf{216.81 $\pm$ 0.87} & \textbf{0.286 $\pm$ 0.001} & \textbf{0.056 $\pm$ 0.003} & \textbf{762.78 $\pm$ 72.39}\phantom{0} \\
\midrule
Direct Code Generation & 3.872 $\pm$ 0.321 & 217.32 $\pm$ 0.75 & 0.288 $\pm$ 0.001 & 0.057 $\pm$ 0.004 & 651.85 $\pm$ 85.93\phantom{0} \\
\bottomrule
\end{tabular}
}
\end{table*}

\paragraph{Additional Examples of Functional Redundancy in Direct Code Generation.}
As shown in \Cref{fig:direct_code_gen}, direct code generation often produces syntactically different implementations that correspond to the same kernel function. 
This is a common pitfall of searching directly in code space: changes such as variable renaming, syntax-level variation, or algebraic rewriting can be counted as new candidates.
In \Cref{fig:dup_kernel_rbfrq,fig:dup_kernel_svm,fig:dup_kernel_recip}, we provide examples of this behavior. 
Across these pairs, the two forward() implementations look different at the code level but share the same kernel function.
\begin{figure}[h]
\centering
\begin{subfigure}[t]{0.48\linewidth}
\begin{lstlisting}[
style=promptstyle,
basicstyle=\ttfamily\scriptsize,
numbers=none,
escapeinside={(*}{*)}
]
def forward(self, x1, x2, diag=False, **params):
    x1s = x1_scaled / self.lengthscale
    x2s = x2_scaled / self.lengthscale
    dist_sq = torch.cdist(x1s, x2s, p=2).pow(2)

    (*\textcolor{varcolor}{covar1}*) = torch.exp(-0.5 * dist_sq).clamp(min=1e-15)
    (*\textcolor{varcolor}{covar2}*) = ((*\textcolor{syntaxcolor}{1.0}*) + dist_sq / ((*\textcolor{syntaxcolor}{2.0}*) * alpha)).pow(-alpha).clamp(min=1e-15)
    covar = covar1 + covar2

    if diag:
        return covar.diagonal(dim1=-2, dim2=-1)
    return covar
\end{lstlisting}
\label{fig:dup_A_rbfrq}
\end{subfigure}
\hfill
\begin{subfigure}[t]{0.48\linewidth}
\begin{lstlisting}[
style=promptstyle,
basicstyle=\ttfamily\scriptsize,
numbers=none,
escapeinside={(*}{*)}
]
def forward(self, x1, x2, diag=False, **params):
    x1s = x1 / self.lengthscale
    x2s = x2 / self.lengthscale
    dist_sq = torch.cdist(x1s, x2s, p=2).pow(2)

    (*\textcolor{varcolor}{covar\_exp}*) = torch.exp(-0.5 * dist_sq).clamp(min=1e-15)
    (*\textcolor{varcolor}{covar\_matern}*) = ((*\textcolor{syntaxcolor}{1}*) + dist_sq / ((*\textcolor{syntaxcolor}{2}*) * self.alpha)).pow(-self.alpha).clamp(min=1e-15)
    covar_combined = covar_exp + covar_matern
    if diag:
        return covar_combined.diagonal(dim1=-2, dim2=-1)
    return covar_combined
\end{lstlisting}
\label{fig:dup_B_rbfrq}
\end{subfigure}
\caption{Functional redundancy example from direct code generation. Both snippets compute \(k(x,x')=\exp(-d^2/2)+(1+d^2/(2\alpha))^{-\alpha}\). \textcolor{varcolor}{Green} denotes variable renaming, and \textcolor{syntaxcolor}{red} denotes syntax-level variation.}
\label{fig:dup_kernel_rbfrq}
\end{figure}

\clearpage
\vspace*{0cm}
\begin{figure}[t]
\centering
\begin{subfigure}[t]{0.48\linewidth}
\begin{lstlisting}[
style=promptstyle,
basicstyle=\ttfamily\scriptsize,
numbers=none,
escapeinside={(*}{*)}
]
def forward(self, x1, x2, diag=False, **params):
    x1s = x1 / self.lengthscale
    x2s = x2 / self.lengthscale
    (*\textcolor{varcolor}{sq\_dist}*) = torch.cdist(x1s, x2s).pow(2)

    covar1 = torch.exp(-0.5 * sq_dist)
    sqrt5_d = math.sqrt(5) * torch.cdist(x1s, x2s).clamp(min=1e-15)


    (*\textcolor{syntaxcolor}{covar2}*) = (1.0 + sqrt5_d + (5.0 / 3.0) * sq_dist) * torch.exp(-sqrt5_d)

    covar = self.alpha * covar1 + (1 - self.alpha) * covar2

    
    if diag:
        return covar.diagonal(dim1=-1, dim2=-2)
    return covar
\end{lstlisting}
\label{fig:dup_A_svm}
\end{subfigure}
\hfill
\begin{subfigure}[t]{0.48\linewidth}
\begin{lstlisting}[
style=promptstyle,
basicstyle=\ttfamily\scriptsize,
numbers=none,
escapeinside={(*}{*)}
]
def forward(self, x1, x2, diag=False, **params):
    x1s = x1 / self.lengthscale
    x2s = x2 / self.lengthscale
    (*\textcolor{varcolor}{dist2}*) = torch.cdist(x1s, x2s).pow(2)

    exp_neg_half_dist2 = torch.exp(-0.5 * dist2)
    scaled_dist = torch.cdist(x1s, x2s).clamp(min=1e-15)
    covar1 = exp_neg_half_dist2
    (*\textcolor{syntaxcolor}{covar2}*) = (1.0 + math.sqrt(5) * scaled_dist + (5.0 / 3.0) * dist2) * torch.exp(-math.sqrt(5) * scaled_dist)

    covar = self.alpha * covar1 + (1 - self.alpha) * covar2

    if diag:
        return covar.diagonal(dim1=-1, dim2=-2)
    return covar
\end{lstlisting}
\label{fig:dup_B_svm}
\end{subfigure}
\caption{Functional redundancy example from direct code generation. Both snippets compute \(k(x,x')=\alpha\exp(-d^2/2)+(1-\alpha)(1+\sqrt{5}d+\frac{5}{3}d^2)\exp(-\sqrt{5}d)\). \textcolor{varcolor}{Green} denotes variable renaming, and \textcolor{syntaxcolor}{red} denotes syntax-level variation.}
\label{fig:dup_kernel_svm}
\end{figure}

\begin{figure}[h]
\centering
\begin{subfigure}[t]{0.48\linewidth}
\begin{lstlisting}[
style=promptstyle,
basicstyle=\ttfamily\scriptsize,
numbers=none,
escapeinside={(*}{*)}
]
def forward(self, x1, x2, diag=False, **params):

    (*\textcolor{syntaxcolor}{x1s}*) = x1 / self.lengthscale
    (*\textcolor{syntaxcolor}{x2s}*) = x2 / self.lengthscale

    dist2 = torch.cdist(x1s, x2s, p=2).pow(2)

    radial_covar = torch.exp(-dist2 / self.alpha).clamp(max=1e15)
    (*\textcolor{algcolor}{negative\_power}*) = (1 + dist2).pow(-self.beta).clamp(min=1e-15)

    (*\textcolor{varcolor}{covar}*) = radial_covar * negative_power


    if diag:
        return covar.diagonal(dim1=-2, dim2=-1)
    return covar
\end{lstlisting}
\label{fig:dup_A_recip}
\end{subfigure}
\hfill
\begin{subfigure}[t]{0.48\linewidth}
\begin{lstlisting}[
style=promptstyle,
basicstyle=\ttfamily\scriptsize,
numbers=none,
escapeinside={(*}{*)}
]
def forward(self, x1, x2, diag=False, **params):
    device = x1.device
    (*\textcolor{syntaxcolor}{x1s}*) = x1 / self.lengthscale.to(device)
    (*\textcolor{syntaxcolor}{x2s}*) = x2 / self.lengthscale.to(device)

    dist2 = torch.cdist(x1s, x2s, p=2).pow(2)

    radial_covar = torch.exp(-dist2 / self.alpha.to(device))
    (*\textcolor{algcolor}{reciprocal\_power}*) = 1.0 / (1.0 + dist2).pow(self.beta.to(device))

    (*\textcolor{varcolor}{combined\_covar}*) = radial_covar * reciprocal_power

    if diag:
        return combined_covar.diagonal(dim1=-2, dim2=-1)
    return combined_covar
\end{lstlisting}
\label{fig:dup_B_recip}
\end{subfigure}
\caption{Functional redundancy example from direct code generation. Both snippets compute \(k(x,x')=\exp(-d^2/\alpha)(1+d^2)^{-\beta}\). \textcolor{varcolor}{Green} denotes variable renaming, \textcolor{syntaxcolor}{red} denotes syntax-level variation, and \textcolor{algcolor}{blue} denotes algebraic rewriting.}
\label{fig:dup_kernel_recip}
\end{figure}



\clearpage
\section{Extended Ablation Studies}\label{app:extend_ablation}
\paragraph{Component Ablation on Other Benchmarks}
In \Cref{fig:system_arch}, we conduct an ablation study on each component of our method on the SVM benchmark. In this section, we summarize the ablation studies on other benchmarks, extending several combinations of our methods. 
As shown in \Cref{tab:ablation}, removing any single component consistently degrades performance. Notably, omitting the Discovery stage or both the Discovery and the Initialization stage results in the most severe performance drops, second only to removing all components. This suggests that these components are crucial design choices for the efficacy of the kernel discovery pipeline in high-dimensional BO.


\begin{table*}[h]
\centering
\caption{Ablation study of Kernel Discovery components across standard benchmarks. $D$ denotes the dimensionality of the task. \textbf{Bold} denotes the best entry in the column. Experiments are conducted with 4 random seeds.}
\label{tab:ablation}
\vspace{2pt}
\renewcommand{\arraystretch}{1.2}

\resizebox{.95\textwidth}{!}{
\begin{tabular}{l cccccc}
\toprule
\multirow{2}{*}{\textbf{Method}} & \textbf{Rover $(\uparrow)$} & \textbf{Mopta08 $(\downarrow)$} & \textbf{Lasso-DNA $(\downarrow)$} & \textbf{SVM388 $(\downarrow)$} & \textbf{Humanoid $(\uparrow)$} & \textbf{Avg Rank $(\downarrow)$} \\
& ($D=100$) & ($D=124$) & ($D=180$) & ($D=388$) & ($D=6392$) & \\
\midrule

\textbf{Kernel Discovery (Ours)} & \textbf{4.353 $\pm$ 0.319} & \textbf{216.81 $\pm$ 0.87} & \textbf{0.286 $\pm$ 0.001} & \textbf{0.056 $\pm$ 0.003} & \textbf{762.78 $\pm$ 72.39}\phantom{0} & \textbf{1.0 / 8} \\
\midrule
w.o. Init & 3.837 $\pm$ 0.390 & 217.60 $\pm$ 2.34 & 0.287 $\pm$ 0.001 & 0.071 $\pm$ 0.011 & 608.40 $\pm$ 63.92\phantom{0} & 4.9 / 8 \\
w.o. Discovery & 3.810 $\pm$ 0.448 & 220.42 $\pm$ 0.95 & 0.288 $\pm$ 0.001 & 0.064 $\pm$ 0.003 & 720.41 $\pm$ 41.77\phantom{0} & 5.2 / 8 \\
w.o. Eval & 3.928 $\pm$ 0.402 & 219.42 $\pm$ 2.37 & 0.289 $\pm$ 0.001 & 0.057 $\pm$ 0.003 & 712.91 $\pm$ 99.62\phantom{0} & 4.0 / 8 \\
w.o. Init + Discovery & 4.153 $\pm$ 0.471 & 217.47 $\pm$ 0.65 & 0.293 $\pm$ 0.003 & 0.079 $\pm$ 0.009 & 557.07 $\pm$ 117.39 & 5.4 / 8 \\
w.o. Init + Eval &  4.132 $\pm$ 0.493 & 218.70 $\pm$ 1.55 & 0.291 $\pm$ 0.004 & 0.065 $\pm$ 0.010 & 610.89 $\pm$ 73.68\phantom{0} & 5.0 / 8 \\
w.o. Discovery + Eval & 3.908 $\pm$ 0.415 & 220.20 $\pm$ 0.59 & 0.287 $\pm$ 0.001 & 0.063 $\pm$ 0.002 & 705.71 $\pm$ 10.51\phantom{0} & 4.3 / 8 \\
w.o. Init + Discovery + Eval & 3.905 $\pm$ 0.321 & 218.56 $\pm$ 1.83 & 0.290 $\pm$ 0.001 & 0.087 $\pm$ 0.009 & 569.78 $\pm$ 88.46\phantom{0} & 6.2 / 8 \\
\bottomrule
\end{tabular}
}
\end{table*}
\begin{figure}[h]
    \centering
    \includegraphics[width=\linewidth]{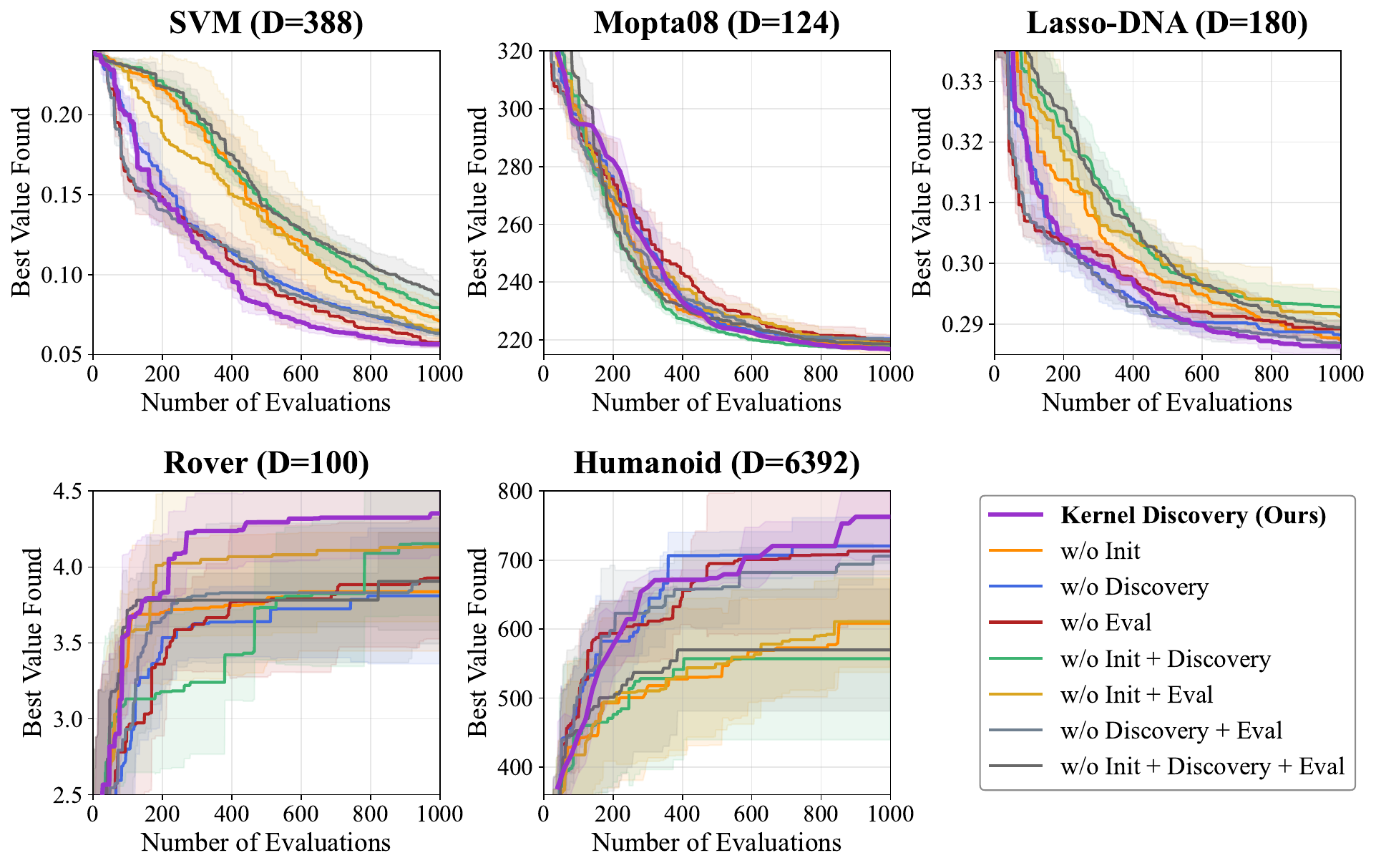}
    \caption{Ablation studies on each component of our method across other benchmarks.}
    \label{fig:system_arch_extended}
    \vspace{-20pt}
\end{figure}

\begin{figure}[h]
    \centering
    \includegraphics[width=\linewidth]{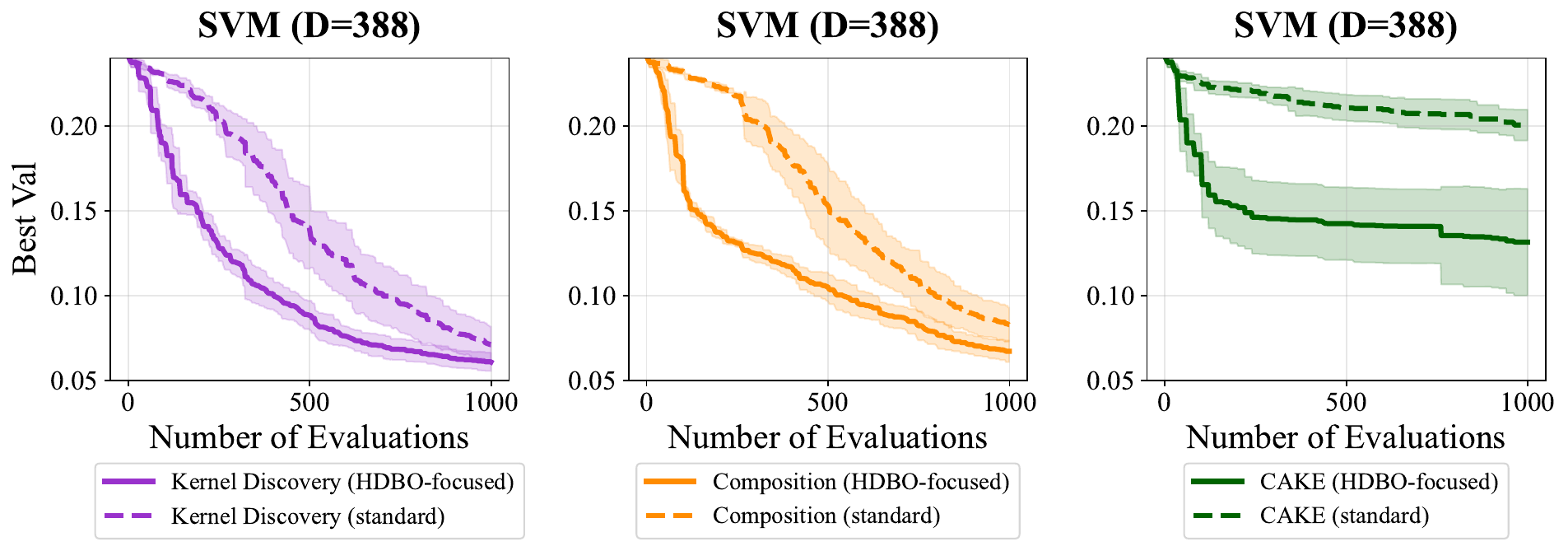}
    \caption{Ablation studies on the base population for Compositional Search and CAKE baselines.}
    \label{fig:base_population_search}
\end{figure}

\clearpage
\paragraph{More Ablations on Base Population.}
For a fair comparison, we also initialize the base populations of the search-based methods, Compositional Search and CAKE, in the same way as in our method. We also conduct ablation studies on the base population for those baselines. As shown in \Cref{fig:base_population_search}, the performance of those methods significantly degrades when we remove BOCK and SL kernels from the base population. While our method also exhibits low sample efficiency without those kernels, it consistently improves the performance through discovering novel kernel structures.

\begin{figure}[h]
    \centering
    \includegraphics[width=\linewidth]{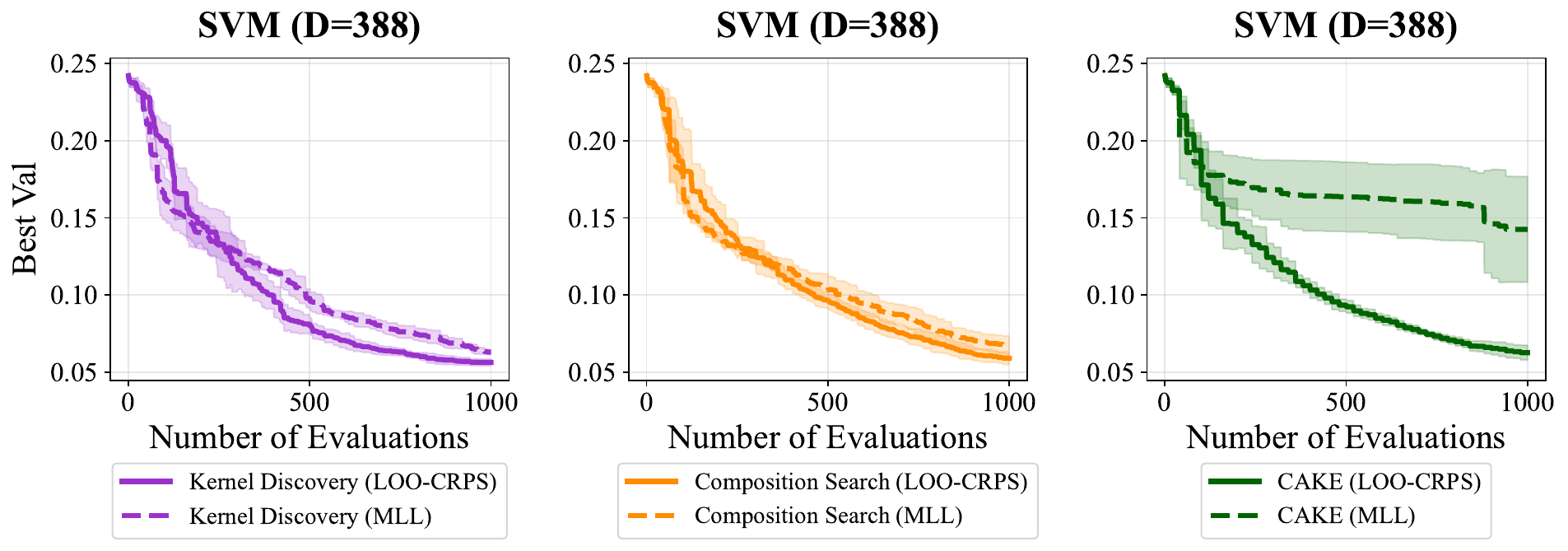}
    \caption{Ablation studies on the evaluation metric for Compositional Search and CAKE baselines.}
    \label{fig:eval_metric_seach}
\end{figure}

\paragraph{More Ablations on Evaluation Metric.}
To follow the original implementation, we use the marginal log-likelihood (MLL) to select a kernel for Compositional Search and BIC-Acquisition Kernel Ranking (BAKER) to select a kernel for CAKE. To analyze the effect of the proposed evaluation metric. We compare performance across different evaluation metrics for both our kernel discovery pipeline and search-based baselines. As shown in \Cref{fig:eval_metric_seach}, performance degrades when we replace the evaluation metric with MLL, which tends to favor overly complex kernels that are likely to overfit to the current dataset. We also observe that LOO-CRPS improves the performance of search-based baselines, indicating that it is a powerful metric for selecting kernels in high-dimensional BO. 

%

\clearpage

\paragraph{Representative Kernel Validation Failures.}
For \(\mathcal{V}_{\mathrm{agn}}\), we test whether a generated implementation satisfies the covariance shape requirements.
GP fitting only requires the self-covariance matrix \(K(\mathbf{X},\mathbf{X})\), but posterior prediction and acquisition evaluation also require the generally rectangular cross-covariance matrix \(K(\mathbf{X}_\ast, \mathbf{X})\).
Accordingly, a common failure mode is that the generated code works for self-covariance calls but breaks under cross-covariance inputs.
\Cref{fig:failure_cases} presents two representative examples.

\begin{figure}[h]
\centering
\begin{subfigure}[t]{0.48\linewidth}
\begin{lstlisting}[
style=promptstyle,
basicstyle=\ttfamily\scriptsize,
numbers=none,
emph={I,ERROR},
emphstyle=\color{red}
]
def forward(self, x1, x2, diag=False, **params):
    # x1: (N1, D), x2: (N2, D)
    # ... input normalization, sphere projection,
    # and tanh warping ...
    geodesic_kernel = ...      # (N1, N2)
    rq_component_TW = ...      # (N1, N2)
    base_kernel = geodesic_kernel * rq_component_TW  # (N1, N2)

    I = torch.eye(x1.size(-2), device=x1.device)[None]  # (1, N1, N1)
    regularized_kernel = base_kernel + self.regularization_strength * I
    # ERROR: shape mismatch when N1 != N2

    if diag:
        return regularized_kernel.diagonal(dim1=-2, dim2=-1)
    return regularized_kernel
\end{lstlisting}
\caption{Identity-matrix regularization. The added \texttt{torch.eye(N1)} creates a square \(N_1 \times N_1\) term, which is only compatible with the covariance tensor shape \(N_1 \times N_2\) when \(N_1 = N_2\).}
\label{fig:failure_eye}
\end{subfigure}
\hfill
\begin{subfigure}[t]{0.48\linewidth}
\begin{lstlisting}[
style=promptstyle,
basicstyle=\ttfamily\scriptsize,
numbers=none,
emph={diagonal,radial_base_kernel,ERROR},
emphstyle=\color{red}
]
def forward(self, x1, x2, diag=False, **params):
    # x1: (N1, D), x2: (N2, D)
    x1 = (x1 - center) / lengthscale
    x2 = (x2 - center) / lengthscale

    dist2 = torch.cdist(x1, x2).square()
    # dist2: (N1, N2)

    r1 = dist2.diagonal(dim1=-2, dim2=-1)
    # r1: (min(N1, N2),)

    angular_kernel = ...
    # angular_kernel: (N1, N2)

    radial_kernel = self.radial_base_kernel(r1.sqrt(), **params)
    # radial_kernel : (min, min)

    return angular_kernel * radial_kernel
    # ERROR: shape mismatch when N1 != N2
\end{lstlisting}
\caption{Implicit square term. The diagonal extraction and sub-kernel call produce an \(M\times M\) block with \(M=\min(N_1,N_2)\), which is incompatible with the \(N_1\times N_2\) angular term.}
\label{fig:failure_diag}
\end{subfigure}
\caption{Representative cross-covariance failures caught by \(\mathcal{V}_{\mathrm{agn}}\). Both implementations pass \(K(\mathbf{X},\mathbf{X})\) but fail on rectangular \(K(\mathbf{X_\ast},\mathbf{X})\) calls. Problematic lines are highlighted in red.}
\label{fig:failure_cases}
\end{figure}

For \(\mathcal{V}_{\mathrm{psd}}\), a generated implementation may satisfy the required output shape but still fail to define a valid covariance function.
A valid GP kernel must produce a positive semi-definite Gram matrix for any finite input set.
A common failure mode is that the generated code adds terms that are not generally PSD-preserving, such as distance-increasing components or oscillatory transformations of pairwise distances.
Such terms can yield indefinite Gram matrices, as illustrated in \Cref{fig:psd_failure_cases}.

\begin{figure}[h]
\centering
\begin{subfigure}[t]{0.48\linewidth}
\begin{lstlisting}[
style=promptstyle,
basicstyle=\ttfamily\scriptsize,
numbers=none,
emph={dist_term},
emphstyle=\color{red}
]
def forward(self, x1, x2, diag=False, **params):
    x1s = x1 / self.lengthscale
    x2s = x2 / self.lengthscale

    dist_sq = torch.cdist(x1s, x2s).pow(2)

    rbf_term = torch.exp(-dist_sq)
    dist_term = torch.sqrt(dist_sq)

    covar = self.alpha * rbf_term + self.beta * dist_term + self.gamma

    if diag:
        return covar.diagonal(dim1=-2, dim2=-1)
    return covar
\end{lstlisting}
\caption{Distance-increasing term. This corresponds to \(k(x,x')=\alpha e^{-d^2}+\beta d+\gamma\). The added distance term \(d=\|x-x'\|\) is not generally PSD-preserving.}\label{fig:failure_psd_dist}
\end{subfigure}
\hfill
\begin{subfigure}[t]{0.48\linewidth}
\begin{lstlisting}[
style=promptstyle,
basicstyle=\ttfamily\scriptsize,
numbers=none,
emph={osc_term},
emphstyle=\color{red}
]
def forward(self, x1, x2, diag=False, **params):
    x1s = x1 / self.lengthscale
    x2s = x2 / self.lengthscale

    dist_sq = torch.cdist(x1s, x2s).pow(2)

    base_term = torch.exp(-0.5 * dist_sq)
    osc_term = torch.cos(math.pi*dist_sq)

    covar = self.alpha * base_term + \
        (1 - self.alpha) * osc_term

    if diag:
        return covar.diagonal(dim1=-2, dim2=-1)
    return covar
\end{lstlisting}
\caption{Oscillatory distance transform. This corresponds to \(k(x,x')=\alpha e^{-d^2/2}+(1-\alpha)\cos(\pi d^2)\). The oscillatory term \(\cos(\pi d^2)\) is not generally PSD-preserving.}\label{fig:failure_psd_cos}
\end{subfigure}
\caption{Representative PSD failures caught by \(\mathcal{V}_{\mathrm{psd}}\). Both include terms that do not generally preserve positive semi-definiteness and can yield indefinite Gram matrices. Problematic terms are highlighted in red.}
\label{fig:psd_failure_cases}
\end{figure}


\clearpage
\section{Robustness to Different LLMs}
\label{app:robustness_llm}

\paragraph{Different LLMs for Other Baselines.}
We additionally evaluate LLM-based baselines under different LLM backbones to examine whether the performance gap is sensitive to the choice of LLM. Specifically, we compare CAKE and LMABO using GPT-4o and GPT-4o-mini, and report the results together with our method in Table~\ref{tab:llm_baselines_4o_mini}. Across the evaluated settings, our method consistently achieves stronger BO performance, suggesting that the performance gain comes from the kernel discovery framework rather than merely from the choice of LLM backbone.

\begin{table*}[h]
\centering
\caption{Performance of LLM-based methods under different LLM backbones.
Results are averaged over 4 random seeds.
Bold denotes the best entry in each column.}
\label{tab:llm_baselines_4o_mini}
\vspace{2pt}
\renewcommand{\arraystretch}{1.2}

\resizebox{.95\textwidth}{!}{
\begin{tabular}{l ccccc}
\toprule
\multirow{2}{*}{\textbf{Method}} & \textbf{Rover $(\uparrow)$} & \textbf{Mopta08 $(\downarrow)$} & \textbf{Lasso-DNA $(\downarrow)$} & \textbf{SVM388 $(\downarrow)$} & \textbf{Humanoid $(\uparrow)$} \\
& ($D=100$) & ($D=124$) & ($D=180$) & ($D=388$) & ($D=6392$) \\
\midrule

\textbf{Kernel Discovery (GPT-4o, Ours)} & \textbf{4.353 $\pm$ 0.319} & \textbf{216.81 $\pm$ 0.87} & \textbf{0.286 $\pm$ 0.001} & \textbf{0.056 $\pm$ 0.003} & \textbf{762.78 $\pm$ 72.39}\phantom{0} \\
Kernel Discovery (GPT-4o-mini, Ours) & 3.582 $\pm$ 0.230 & 219.16 $\pm$ 1.54\phantom{0} & 0.289 $\pm$ 0.002 & 0.061 $\pm$ 0.004 & 589.28 $\pm$ 107.87 \\
\midrule

CAKE (GPT-4o) & 3.412 $\pm$ 0.586 & 231.40 $\pm$ 15.67 & 0.300 $\pm$ 0.011 & 0.131 $\pm$ 0.036 & 667.49 $\pm$ 24.99\phantom{0} \\
CAKE (GPT-4o-mini) & 3.863 $\pm$ 0.497 & 229.43 $\pm$ 3.88\phantom{0} & 0.290 $\pm$ 0.001 & 0.103 $\pm$ 0.058 & 648.13 $\pm$ 29.11\phantom{0} \\
\midrule

LMABO (GPT-4o) & 3.953 $\pm$ 0.369 & 240.70 $\pm$ 5.24\phantom{0} & 0.304 $\pm$ 0.006 & 0.226 $\pm$ 0.001 & 390.23 $\pm$ 44.78\phantom{0} \\
LMABO (GPT-4o-mini) & 4.078 $\pm$ 0.299 & 244.21 $\pm$ 3.97\phantom{0} & 0.302 $\pm$ 0.005 & 0.218 $\pm$ 0.016 & 410.09 $\pm$ 42.34\phantom{0} \\
\bottomrule
\end{tabular}
}
\end{table*}

\paragraph{Open-source LLMs for Kernel Discovery.}
We further evaluate whether our kernel discovery pipeline can operate with a smaller open-source LLM. To this end, we replace the kernel-proposing LLM with Qwen3-8B while keeping the rest of the pipeline unchanged. As shown in \Cref{tab:open_source_llm}, GPT-4o achieves the best performance, but the Qwen3-8B variant still obtains competitive results across benchmarks. Compared with the main results in \Cref{tab:benchmark_results}, the open-source variant remains comparable to strong baseline methods, indicating that our framework does not rely exclusively on a proprietary LLM. These results suggest that stronger LLMs improve the quality of discovered kernels, while the proposed pipeline remains effective even with a relatively small open-source model.

\begin{table*}[h]
\centering
\caption{Performance of Kernel Discovery with proprietary and open-source LLMs.
Results are averaged over 4 random seeds.
Bold denotes the best entry in each column.}
\label{tab:open_source_llm}
\vspace{2pt}
\renewcommand{\arraystretch}{1.2}

\resizebox{.95\textwidth}{!}{
\begin{tabular}{l ccccc}
\toprule
\multirow{2}{*}{\textbf{Method}} & \textbf{Rover $(\uparrow)$} & \textbf{Mopta08 $(\downarrow)$} & \textbf{Lasso-DNA $(\downarrow)$} & \textbf{SVM388 $(\downarrow)$} & \textbf{Humanoid $(\uparrow)$} \\
& ($D=100$) & ($D=124$) & ($D=180$) & ($D=388$) & ($D=6392$) \\
\midrule

\textbf{Kernel Discovery (GPT-4o, Ours)} & \textbf{4.353 $\pm$ 0.319} & \textbf{216.81 $\pm$ 0.87} & \textbf{0.286 $\pm$ 0.001} & \textbf{0.056 $\pm$ 0.003} & \textbf{762.78 $\pm$ 72.39}\phantom{00} \\
Kernel Discovery (Qwen3-8B, Ours) & 3.838 $\pm$ 0.293 & 221.67 $\pm$ 1.53 & 0.288 $\pm$ 0.002 & 0.058 $\pm$ 0.003 & 622.15 $\pm$ 107.11 \\
\bottomrule
\end{tabular}
}
\end{table*}

%

\clearpage
\section{Evolution of Discovered Kernels}
\label{app:insights}
\paragraph{Detailed Explanation of Discovered Kernels in Figure~\ref{fig:insights}.}
We provide an explanation of the discovered kernels highlighted in the figure. 
We first define the kernel components used in the discovered kernels, and then describe each discovered kernel by its decomposition and its interpretations.

\emph{Kernel components.}
We first define the kernel components that appear in the discovered kernels. 
For the RQ component evaluated in the original input space, let $\boldsymbol{\ell}_B\in\mathbb{R}_{>0}^d$ be an ARD lengthscale, $r_{\mathrm{RQ}} = \|\mathbf{x}\oslash\boldsymbol{\ell}_B - \mathbf{x}'\oslash\boldsymbol{\ell}_B\|_2$, and $\alpha>0$.
Then
\[
k_{\mathrm{RQ}}(\mathbf{x},\mathbf{x}')
=
\left(
1+
\frac{r_{\mathrm{RQ}}^2}{2\alpha}
\right)^{-\alpha}.
\]
This component captures multi-scale smooth variation in a separately ARD-scaled input space.

Several other components are evaluated after an arctangent-based input transformation. 
Given an input $\mathbf{x}\in\mathbb{R}^d$ and an ARD lengthscale $\boldsymbol{\ell}\in\mathbb{R}_{>0}^d$, define
\[
\mathbf{z}_0(\mathbf{x})=\mathbf{x}\oslash\boldsymbol{\ell},
\qquad
\mathbf{z}_i(\mathbf{x})
=
\frac{1}{\sqrt{i}}
\arctan\!\left(s w\,\mathbf{z}_{i-1}(\mathbf{x})\right),
\quad i=1,\dots,D,
\]
where $s,w>0$ are scalar transformation parameters and $D\in\mathbb{N}$ is the number of transformation layers. 
We denote the final transformed feature by $t(\mathbf{x})=\mathbf{z}_D(\mathbf{x}).$
Using this transformed feature, we define
\[
k_{\mathrm{Arc\text{-}IMQ}}(\mathbf{x},\mathbf{x}')
=
\left(
1+\|t(\mathbf{x})-t(\mathbf{x}')\|_2^2
\right)^{-1},\quad
k_{\mathrm{Arc\text{-}Lin}}(\mathbf{x},\mathbf{x}')
=
t(\mathbf{x})^\top t(\mathbf{x}'),
\]
and
\[
k_{\mathrm{Arc\text{-}RQ}}(\mathbf{x},\mathbf{x}')
=
\left(
1+
\frac{\|t(\mathbf{x})-t(\mathbf{x}')\|_2^2}{2\alpha}
\right)^{-\alpha}.
\]
Here, the Arc-IMQ component captures heavy-tailed distance-based similarity in the arctangent-transformed feature space, the Arc-Linear component captures global alignment in the same transformed space, and the Arc-RQ component captures multi-scale smooth variation after the arctangent transformation.

One discovered kernel additionally uses an angular similarity component. 
Let
\[
\tilde{\mathbf{x}}
=
\frac{\mathbf{x}-\mathbf{c}}{\mathbf{r}},
\qquad
\rho(\mathbf{x})
=
\|\tilde{\mathbf{x}}\|_2,
\qquad
\mathbf{a}(\mathbf{x})
=
\frac{\tilde{\mathbf{x}}}{\rho(\mathbf{x})}.
\]
Here $\mathbf{a}(\mathbf{x})$ is the unit direction vector of the normalized input. 
The polynomial angular kernel is
\[
k_{\mathrm{ang}}(\mathbf{x},\mathbf{x}')
=
\sum_{p=0}^{3}
w_p
\left(
\mathbf{a}(\mathbf{x})^\top
\mathbf{a}(\mathbf{x}')
\right)^p,
\qquad w_p>0.
\]
This component captures directional similarity relative to the normalized center.

\emph{Kernel 1:}
\[
k_1(\mathbf{x},\mathbf{x}')
=
k_{\mathrm{Arc\text{-}IMQ}}(\mathbf{x},\mathbf{x}')
+
k_{\mathrm{Arc\text{-}Lin}}(\mathbf{x},\mathbf{x}')
+
k_{\mathrm{RQ}}(\mathbf{x},\mathbf{x}').
\]
This kernel combines heavy-tailed distance-based similarity and linear alignment in the arctangent-transformed feature space with multi-scale smooth variation in a separate ARD-scaled input space.

\emph{Kernel 2:}
\[
k_2(\mathbf{x},\mathbf{x}')
=
k_{\mathrm{ang}}(\mathbf{x},\mathbf{x}')
\cdot
\left[
k_{\mathrm{Arc\text{-}IMQ}}(\mathbf{x},\mathbf{x}')
+
k_{\mathrm{Arc\text{-}Lin}}(\mathbf{x},\mathbf{x}')
\right].
\]
This kernel is a product of a polynomial angular kernel and an arc-transformed additive kernel. 
The angular component captures directional similarity relative to a normalized center, while the arc component captures heavy-tailed distance-based similarity and linear alignment in the arctangent-transformed feature space.

\emph{Kernel 3:}
\[
k_3(\mathbf{x},\mathbf{x}')
=
k_{\mathrm{Arc\text{-}RQ}}(\mathbf{x},\mathbf{x}')
+
k_{\mathrm{Arc\text{-}IMQ}}(\mathbf{x},\mathbf{x}')
+
k_{\mathrm{Arc\text{-}Lin}}(\mathbf{x},\mathbf{x}').
\]
This kernel evaluates all three components in the same arctangent-transformed feature space. 
It combines multi-scale smooth variation, heavy-tailed distance-based similarity, and global transformed-space alignment under a shared nonlinear input warping.

We also provide the corresponding \texttt{forward} implementations for the discovered kernels.

\begin{figure}[h!]
\begin{nolinenumbers}
\begin{lstlisting}[style=promptstyle, caption={Forward code for the discovered kernel \(k_1\).}, label={fig:discovered_code_svm_1}]
def forward(self, x1, x2, diag=False, **params):
    if x1.dim() == 1: x1 = x1.unsqueeze(-1)
    if x2.dim() == 1: x2 = x2.unsqueeze(-1)

    # Kernel A: Arc-transformed IMQ + Arc-transformed Linear
    x1_scaled_a = x1 / self.lengthscale_a
    x2_scaled_a = x2 / self.lengthscale_a
    t1, t2 = x1_scaled_a, x2_scaled_a
    for i in range(1, self.depth.item() + 1):
        t1 = torch.atan(self.global_scale * self.arc_weight * t1) / math.sqrt(i)
        t2 = torch.atan(self.global_scale * self.arc_weight * t2) / math.sqrt(i)
    linear_term = t1 @ t2.transpose(-1, -2)
    r_squared_a = torch.cdist(t1, t2, p=2).pow(2)
    k_a = (1 + r_squared_a).pow(-1) + linear_term

    # Kernel B: Rational Quadratic
    x1_scaled_b = x1 / self.lengthscale_b
    x2_scaled_b = x2 / self.lengthscale_b
    r_squared_b = torch.cdist(x1_scaled_b, x2_scaled_b, p=2).pow(2)
    k_b = (1 + r_squared_b / (2 * self.alpha)).pow(-self.alpha)

    covar = k_a + k_b
    if diag:
        return covar.diagonal(dim1=-2, dim2=-1)
    return covar
\end{lstlisting}
\end{nolinenumbers}
\end{figure}

\begin{figure}[h!]
\begin{nolinenumbers}
\begin{lstlisting}[style=promptstyle, caption={Forward code for the discovered kernel \(k_2\).}, label={fig:discovered_code_svm_box4}]
def forward(self, x1, x2, diag=False, **params):
    if x1.dim() == 1: x1 = x1.unsqueeze(-1)
    if x2.dim() == 1: x2 = x2.unsqueeze(-1)
    arc_weight = self.raw_arc_weight_constraint.transform(self.raw_arc_weight)
    global_scale = self.raw_global_scale_constraint.transform(self.raw_global_scale)
    angular_weights = self.raw_angular_weights_constraint.transform(self.raw_angular_weights)

    x1_normalized = (x1 - self.center) / self.radius
    x2_normalized = (x2 - self.center) / self.radius
    r1 = x1_normalized.norm(dim=-1, keepdim=True)
    r2 = x2_normalized.norm(dim=-1, keepdim=True)
    a1 = x1_normalized / r1.clamp(min=1e-15)
    a2 = x2_normalized / r2.clamp(min=1e-15)

    t1 = x1_normalized / self.lengthscale
    t2 = x2_normalized / self.lengthscale
    for i in range(1, self.depth + 1):
        t1 = torch.atan(global_scale * arc_weight * t1) / math.sqrt(i)
        t2 = torch.atan(global_scale * arc_weight * t2) / math.sqrt(i)

    angular_cov = self._angular_kernel(a1, a2, angular_weights)
    t1_dist = torch.cdist(t1, t2, p=2)
    arc_cov = (1 + t1_dist.pow(2)).pow(-1) + (t1 @ t2.transpose(-1, -2))

    covar = angular_cov * arc_cov
    if diag:
        return covar.diagonal(dim1=-2, dim2=-1)
    return covar
\end{lstlisting}
\end{nolinenumbers}
\end{figure}

\begin{figure}[h!]
\begin{nolinenumbers}
\begin{lstlisting}[style=promptstyle, caption={Forward code for the discovered kernel \(k_3\).}, label={fig:discovered_code_svm}]
def forward(self, x1, x2, diag=False, **params):
    if x1.dim() == 1: x1 = x1.unsqueeze(-1)
    if x2.dim() == 1: x2 = x2.unsqueeze(-1)

    # Kernel A: Arc-transformed IMQ + Arc-transformed Linear
    x1_scaled_a = x1 / self.lengthscale_a
    x2_scaled_a = x2 / self.lengthscale_a
    t1, t2 = x1_scaled_a, x2_scaled_a
    for i in range(1, self.depth.item() + 1):
        t1 = torch.atan(self.global_scale * self.arc_weight * t1) / math.sqrt(i)
        t2 = torch.atan(self.global_scale * self.arc_weight * t2) / math.sqrt(i)
    linear_term = t1 @ t2.transpose(-1, -2)
    r_squared_a = torch.cdist(t1, t2, p=2).pow(2)
    k_a = (1 + r_squared_a).pow(-1) + linear_term

    # Kernel B: Rational Quadratic
    x1_scaled_b = x1 / self.lengthscale_b
    x2_scaled_b = x2 / self.lengthscale_b
    r_squared_b = torch.cdist(x1_scaled_b, x2_scaled_b, p=2).pow(2)
    k_b = (1 + r_squared_b / (2 * self.alpha)).pow(-self.alpha)

    covar = k_a + k_b
    if diag:
        return covar.diagonal(dim1=-2, dim2=-1)
    return covar
\end{lstlisting}
\end{nolinenumbers}
\vspace{-20pt}
\end{figure}



\begin{figure}[h!]
\centering
\includegraphics[width=0.9\linewidth]{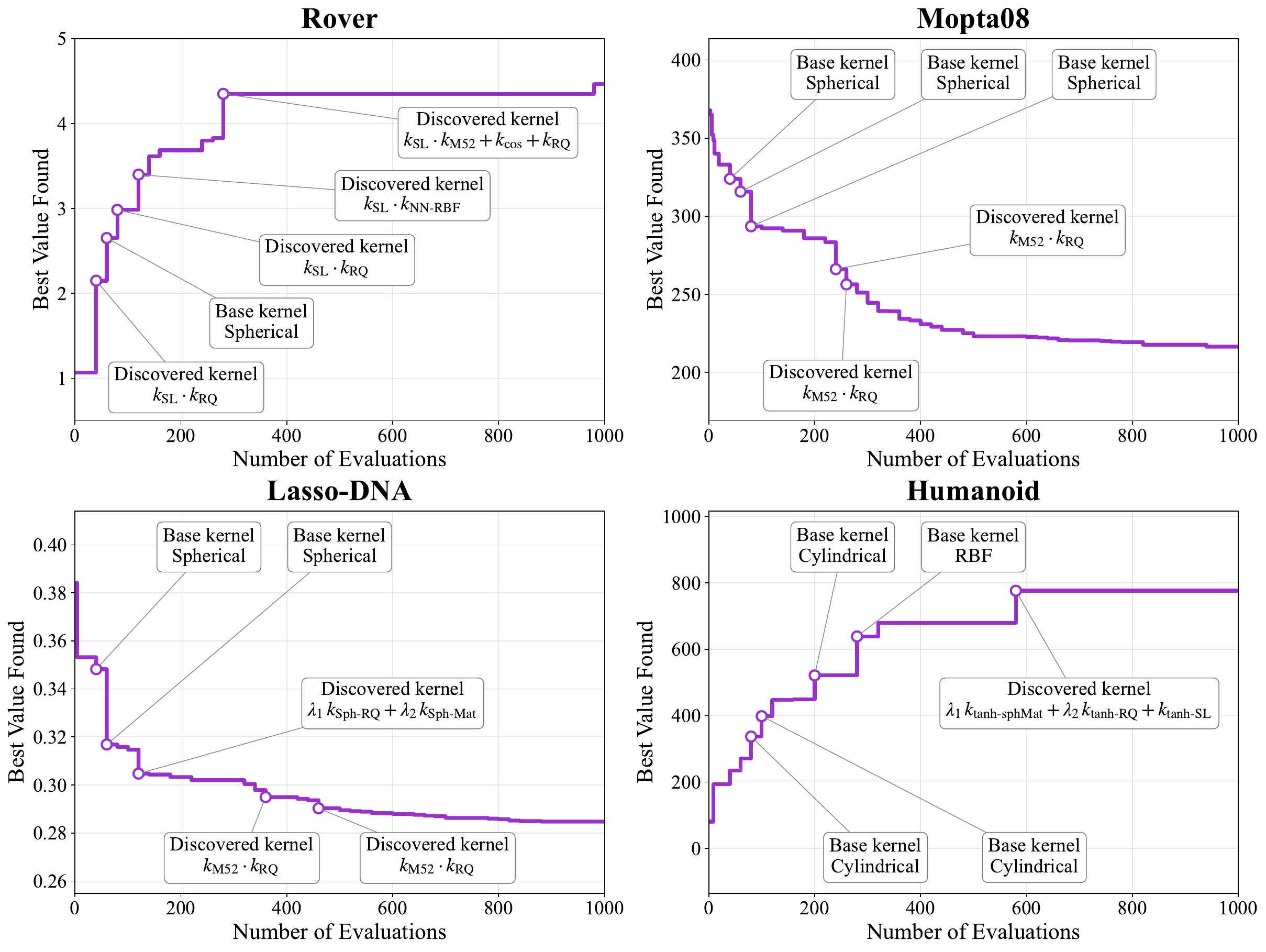}
\caption{
Post-analysis of kernel discovery on four additional benchmarks. 
For each benchmark, we annotate the kernels selected at the top-5 largest improvements in the best-so-far value.
}
\label{fig:insights_other_benchmarks}
\vspace{-20pt}
\end{figure}

\paragraph{Evolving Population on Other Benchmarks}
\Cref{fig:insights_other_benchmarks} extends the post-analysis of \Cref{fig:insights} to four additional benchmarks: Rover, Mopta08, Lasso-DNA, and Humanoid. 
For each benchmark, we visualize the best-so-far value trajectory and annotate the kernels selected at the top-5 largest improvements. 
The components of the annotated kernels are described using the same kernel vocabulary introduced above.

\clearpage

For the additional discovered kernels annotated in Figure~\ref{fig:insights_other_benchmarks}, we provide the mathematical definitions of the component kernels used in the decompositions. 

\emph{Matérn-\(5/2\) component.} 
Let
\[
r_{\text{Matérn52}}
=
\left\|
\mathbf{x}\oslash\boldsymbol{\ell}_{\text{Matérn52}}
-
\mathbf{x}'\oslash\boldsymbol{\ell}_{\text{Matérn52}}
\right\|_2.
\]
Then
\[
k_{\text{Matérn52}}(\mathbf{x},\mathbf{x}')
=
\left(
1+\sqrt{5}r_{\text{Matérn52}}
+
\frac{5}{3}r_{\text{Matérn52}}^2
\right)
\exp(-\sqrt{5}r_{\text{Matérn52}}).
\]

\emph{Spherical lifted linear component.}
Given a scaled input \(\mathbf{u}(\mathbf{x})\), define the stereographic projection
\[
\psi(\mathbf{u}(\mathbf{x}))
=
\frac{
\left[
2\mathbf{u}(\mathbf{x}),\;
\|\mathbf{u}(\mathbf{x})\|_2^2-1
\right]
}{
\|\mathbf{u}(\mathbf{x})\|_2^2+1
}.
\]
Then
\[
k_{\mathrm{SL}}(\mathbf{x},\mathbf{x}')
=
\lambda_0
+
\lambda_1
\psi(\mathbf{u}(\mathbf{x}))^\top
\psi(\mathbf{u}(\mathbf{x}')),
\qquad
\lambda_0+\lambda_1=1,\;\lambda_0,\lambda_1\geq 0.
\]

\emph{Neural-network RBF component.}
Let
\[
\mathbf{h}(\mathbf{x})
=
\operatorname{NN}\!\left(
\mathbf{x}\oslash\boldsymbol{\ell}_{\mathrm{NN}}
\right).
\]
Then
\[
k_{\mathrm{NN\text{-}RBF}}(\mathbf{x},\mathbf{x}')
=
\exp\left(
-
\frac{
\|\mathbf{h}(\mathbf{x})-\mathbf{h}(\mathbf{x}')\|_2^2
}{
2\sigma_{\mathrm{NN}}^2
}
\right).
\]

\emph{Cosine warping component.}
Define
\[
\mathbf{v}(\mathbf{x})
=
\tanh(\mathbf{x}/r_{\mathrm{cos}}),
\qquad
\rho_{\mathrm{cos}}(\mathbf{x})
=
\|\mathbf{v}(\mathbf{x})\|_2.
\]
Then
\[
k_{\mathrm{cos}}(\mathbf{x},\mathbf{x}')
=
\cos\left(
2\pi\alpha_{\mathrm{cos}}
\left|
\rho_{\mathrm{cos}}(\mathbf{x})
-
\rho_{\mathrm{cos}}(\mathbf{x}')
\right|
\right).
\]

\emph{Spherical RQ and spherical Matérn-like components.}
Using the stereographic projection \(\psi(\mathbf{u}(\mathbf{x}))\), define
\[
d_{\mathrm{sph}}
=
\|\psi(\mathbf{u}(\mathbf{x}))-\psi(\mathbf{u}(\mathbf{x}'))\|_2.
\]
The spherical RQ component is
\[
k_{\mathrm{sphRQ}}(\mathbf{x},\mathbf{x}')
=
\left(
1+
\frac{d_{\mathrm{sph}}^2}{2\alpha}
\right)^{-\alpha},
\]
and the spherical Matérn-like component is
\[
k_{\mathrm{sphMat}}(\mathbf{x},\mathbf{x}')
=
\left(
1+
\frac{\sqrt{3}}{\nu}d_{\mathrm{sph}}
+
\frac{3}{\nu^2}d_{\mathrm{sph}}^2
\right)
\exp\left(
-
\frac{\sqrt{3}}{\nu}d_{\mathrm{sph}}
\right).
\]
where \(\nu>0\) is a learnable parameter that controls the effective length scale and smoothness of the Matérn-like component.

\emph{Tanh-transformed components.}
Define the tanh embedding
\[
\mathbf{h}_{\tanh}(\mathbf{x})
=
\tanh\left(
\alpha(\mathbf{x}\oslash\boldsymbol{\ell})
\right).
\]
For the spherical branch, define
\[
\mathbf{u}_{\tanh}(\mathbf{x})
=
(\mathbf{h}_{\tanh}(\mathbf{x})-\mathbf{c})
\oslash
\boldsymbol{\ell},
\]
and let \(\psi_{\tanh}(\mathbf{x}) = \psi(\mathbf{u}_{\tanh}(\mathbf{x}))\) be the stereographic projection of \(\mathbf{u}_{\tanh}(\mathbf{x})\). Define
\[
r_{\tanh\text{-}\mathrm{sph}}
=
\|\psi_{\tanh}(\mathbf{x})-\psi_{\tanh}(\mathbf{x}')\|_2.
\]
Then
\[
k_{\mathrm{tanh\text{-}sphMat}}(\mathbf{x},\mathbf{x}')
=
\left(
1+\sqrt{2\nu}r_{\tanh\text{-}\mathrm{sph}}
+
\frac{2\nu}{3}r_{\tanh\text{-}\mathrm{sph}}^2
\right)
\exp\left(
-\sqrt{2\nu}r_{\tanh\text{-}\mathrm{sph}}
\right),
\]
\[
k_{\mathrm{tanh\text{-}RQ}}(\mathbf{x},\mathbf{x}')
=
\left(
1+
\frac{
\|\mathbf{h}_{\tanh}(\mathbf{x})-\mathbf{h}_{\tanh}(\mathbf{x}')\|_2^2
}{
2\gamma
}
\right)^{-\gamma},
\]
and
\[
k_{\mathrm{tanh\text{-}SL}}(\mathbf{x},\mathbf{x}')
=
\psi_{\tanh}(\mathbf{x})^\top
\psi_{\tanh}(\mathbf{x}').
\]

\clearpage
\section{Transferability of Discovered Kernels}
\label{app:transferability}

\paragraph{Experiment Results on Other Benchmarks.}
In \Cref{fig:transferability}, we visualize the results of the kernel discovered by our pipeline on the SVM benchmark. In this section, we report the kernel's results across the other benchmarks. As shown in \Cref{tab:transferability_extended}, it consistently achieves competitive and often outperforms other base kernels. We also visualize the code snippet of the kernel in \Cref{fig:discovered_code}.

\begin{table*}[h]
\centering
\caption{Transferability of the discovered kernel across other benchmarks. We compare the kernel discovered on the SVM benchmark, when transferred to the other four standard benchmarks, against base kernels. $D$ denotes the dimensionality of the task. \textbf{\textcolor{blue}{Blue}} denotes the best entry in the column, and \textbf{\textcolor{violet}{Violet}} denotes the second best. Experiments are conducted with 4 random seeds.}
\label{tab:transferability_extended}
\vspace{2pt}
\renewcommand{\arraystretch}{1.2}

\resizebox{\textwidth}{!}{
\begin{tabular}{l ccccc c}
\toprule
\multirow{2}{*}{\textbf{Method}} & \textbf{Rover $(\uparrow)$} & \textbf{Mopta08 $(\downarrow)$} & \textbf{Lasso-DNA $(\downarrow)$} & \textbf{SVM $(\downarrow)$} & \textbf{Humanoid $(\uparrow)$} & \textbf{Average} \\
& ($D=100$) & ($D=124$) & ($D=180$) & ($D=388$) & ($D=6392$) & \textbf{Rank} \\
\midrule

\multicolumn{7}{l}{\textbf{Base Kernels}} \\
\midrule
RBF & 3.552 $\pm$ 0.304 & 217.75 $\pm$ 2.33 & \textbf{\textcolor{blue}{0.291 $\pm$ 0.001}} & \textbf{\textcolor{violet}{0.061 $\pm$ 0.004}} & 503.00 $\pm$ 67.82\phantom{0} & \phantom{0}3.8 / 8 \\
Matérn52 & 3.453 $\pm$ 0.366 & \textbf{\textcolor{blue}{215.77 $\pm$ 0.57}} & 0.292 $\pm$ 0.003 & 0.063 $\pm$ 0.003 & 509.79 $\pm$ 102.32 & \phantom{0}4.0 / 8 \\
Linear & 3.950 $\pm$ 0.441 & 274.67 $\pm$ 8.87 & 0.312 $\pm$ 0.004 & 0.226 $\pm$ 0.003 & 435.49 $\pm$ 42.00\phantom{0} & \phantom{0}6.4 / 8 \\
Periodic & 3.664 $\pm$ 0.231 & 309.10 $\pm$ 5.01 & 0.332 $\pm$ 0.005 & 0.226 $\pm$ 0.002 & 413.83 $\pm$ 71.30\phantom{0} & \phantom{0}7.4 / 8 \\
BOCK & 3.725 $\pm$ 0.593 & 225.36 $\pm$ 1.82 & \textbf{\textcolor{violet}{0.291 $\pm$ 0.003}} & 0.068 $\pm$ 0.007 & \textbf{\textcolor{blue}{669.52 $\pm$ 50.15}}\phantom{0} & \phantom{0}\textbf{\textcolor{violet}{3.4 / 8}} \\
SL & \textbf{\textcolor{violet}{4.096 $\pm$ 0.473}} & 246.78 $\pm$ 3.92 & 0.297 $\pm$ 0.001 & 0.112 $\pm$ 0.007 & \textbf{\textcolor{violet}{637.72 $\pm$ 37.86}}\phantom{0} & \phantom{0}4.4 / 8 \\
RQ & 3.858 $\pm$ 0.515 & \textbf{\textcolor{violet}{217.53 $\pm$ 4.25}} & 0.294 $\pm$ 0.002 & 0.069 $\pm$ 0.017 & 600.80 $\pm$ 152.49 & \phantom{0}3.8 / 8 \\
\midrule

\textbf{Discovered Kernel (Ours)} & \textbf{\textcolor{blue}{4.238 $\pm$ 0.640}} & 218.30 $\pm$ 1.00 & 0.296 $\pm$ 0.004 & \textbf{\textcolor{blue}{0.052 $\pm$ 0.006}} & 629.12 $\pm$ 122.84 & \phantom{0}\textbf{\textcolor{blue}{2.8 / 8}} \\
\bottomrule
\end{tabular}
}
\end{table*}

\begin{figure}[h]
    \centering
    \includegraphics[width=\linewidth]{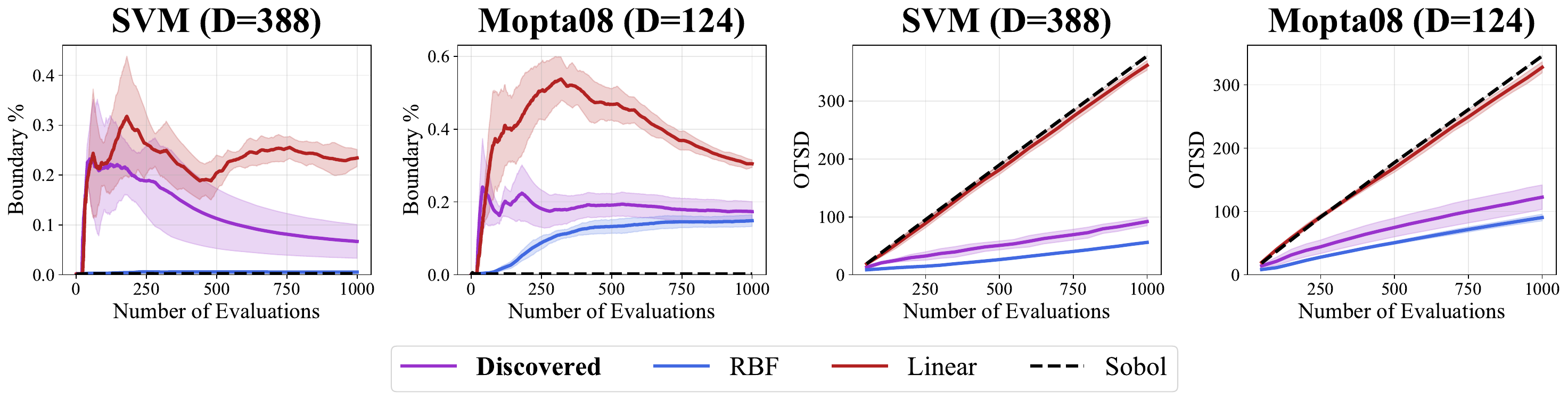}
    \caption{Analysis on boundary-seeking behavior of several kernels.}
    \label{fig:boundary_seeking}
\end{figure}

\paragraph{Analysis on Boundary-seeking Behavior.} To understand why the discovered kernel $k_{\text{discover}}$ in \Cref{eq:discover} achieves superior performance on the SVM benchmark, we analyze whether its non-stationary component $k_{\text{tanh\text{-}Poly}}$ induces boundary-seeking behavior. We measure two complementary statistics over BO iterations: (1) the boundary hit ratio, i.e., the fraction of queried points lying on the boundary of the search domain, and (2) the observation traveling salesman distance (OTSD), which measures the total path length traversed across sequential queries. As shown in \Cref{fig:boundary_seeking}, the discovered kernel maintains a boundary hit ratio comparable to RBF throughout optimization, and substantially lower than the Linear kernel, on both the SVM and Mopta08 benchmarks. The OTSD of the discovered kernel similarly remains low and closely tracks that of RBF, indicating that consecutive queries tend to be spatially concentrated rather than widely scattered. Together, these results suggest that the non-stationary component is regularized during GP hyperparameter optimization, and that the combination of several kernel components encourages a locally concentrated query behavior, which is a key factor enabling the discovered kernel to consistently identify higher-quality candidates.

\paragraph{Detailed Explanation of the Discovered Kernel in \Cref{eq:discover}.}
We provide a detailed explanation of the discovered kernel in \Cref{eq:discover}. 
This kernel combines a Matérn-\(5/2\) component with an additive composition of a tanh-polynomial component and an RQ component:
\[
k_{\mathrm{discover}}(\mathbf{x},\mathbf{x}')
=
k_{\text{Matérn52}}(\mathbf{x},\mathbf{x}')
\cdot
\left[
k_{\mathrm{tanh\text{-}Poly}}(\mathbf{x},\mathbf{x}')
+
k_{\mathrm{RQ}}(\mathbf{x},\mathbf{x}')
\right].
\]

\emph{Matérn-\(5/2\) component.}
Let \(\boldsymbol{\ell}\in\mathbb{R}_{>0}^d\) be an ARD lengthscale and define
\[
r_{\text{Matérn52}}(\mathbf{x},\mathbf{x}')
=
\left\|
\mathbf{x}\oslash\boldsymbol{\ell}
-
\mathbf{x}'\oslash\boldsymbol{\ell}
\right\|_2.
\]
The Matérn-\(5/2\) component is
\[
k_{\text{Matérn52}}(\mathbf{x},\mathbf{x}')
=
\left(
1+\sqrt{5}r_{\text{Matérn52}}
+
\frac{5}{3}r_{\text{Matérn52}}^2
\right)
\exp\left(
-\sqrt{5}r_{\text{Matérn52}}
\right).
\]
This component captures moderately smooth local correlation in the ARD-scaled input space.

\emph{Tanh-polynomial component.}
Using the same ARD-scaled input, define the tanh-transformed feature
\[
\mathbf{h}_{\tanh}(\mathbf{x})
=
\tanh\left(
\sigma
\left(\mathbf{x}\oslash\boldsymbol{\ell}\right)
\right),
\]
where \(\sigma>0\) controls the strength of the tanh projection. 
The tanh-polynomial component is
\[
k_{\mathrm{tanh\text{-}Poly}}(\mathbf{x},\mathbf{x}')
=
\sum_{n=0}^{2}
w_n
\left(
\mathbf{h}_{\tanh}(\mathbf{x})^\top
\mathbf{h}_{\tanh}(\mathbf{x}')
\right)^n,
\qquad w_n>0.
\]
This component captures polynomial similarity after mapping inputs into a bounded nonlinear feature space.

\emph{Rational Quadratic component.}
Let
\[
r_{\mathrm{RQ}}(\mathbf{x},\mathbf{x}')
=
\left\|
\mathbf{x}\oslash\boldsymbol{\ell}
-
\mathbf{x}'\oslash\boldsymbol{\ell}
\right\|_2.
\]
The RQ component is
\[
k_{\mathrm{RQ}}(\mathbf{x},\mathbf{x}')
=
\left(
1+
\frac{r_{\mathrm{RQ}}^2}{2\alpha}
\right)^{-\alpha},
\qquad \alpha>0.
\]
This component captures multi-scale smooth variation in the ARD-scaled input space.

Overall, the product structure modulates the additive tanh-polynomial and RQ similarities by the Matérn-\(5/2\) local smoothness component.

\paragraph{Illustrative Example of Discovered Kernel.} 
As a representative example, we present the full implementation of the kernel discovered from the SVM benchmark, which corresponds to $k_{\text{discover}}$ in \Cref{eq:discover}.

\begin{nolinenumbers}
\begin{lstlisting}[style=promptstyle, caption={Code snippet of the discovered kernel from SVM benchmark.}, label={fig:discovered_code}]
def forward(self, x1, x2, diag=False, **params):
    if x1.dim() == 1: x1 = x1.unsqueeze(-1)
    if x2.dim() == 1: x2 = x2.unsqueeze(-1)
    x1s = x1 / self.lengthscale
    x2s = x2 / self.lengthscale

    # Matern-5/2 Component
    r = torch.cdist(x1s, x2s, p=2).clamp(min=1e-15)
    sqrt5_r = math.sqrt(5.0) * r
    k_matern = (1 + sqrt5_r + 5.0/3.0 * r.pow(2)) \
                * torch.exp(-sqrt5_r)

    # Tangential Polynomial Component
    t1 = torch.tanh(self.projection_sigma * x1s)
    t2 = torch.tanh(self.projection_sigma * x2s)
    t_ip = t1 @ t2.transpose(-1, -2)
    k_tanh_poly = sum(
        self.tangent_weights[n] * t_ip.pow(n)
        for n in range(3))

    # Rational Quadratic Component
    r2 = torch.cdist(x1s, x2s, p=2).pow(2)
    k_rq = (1 + r2 / (2 * self.alpha)).pow(-self.alpha)

    covar = k_matern * (k_tanh_poly + k_rq)
    if diag:
        return covar.diagonal(dim1=-2, dim2=-1)
    return covar
\end{lstlisting}
\end{nolinenumbers}

We additionally present two discovered kernels that show strong transferability. Using the same fixed-kernel evaluation protocol as in \Cref{tab:transferability_extended}, both achieve an overall average rank of 2.8 across five benchmarks, outperforming all standard base kernels in aggregate.

\begin{nolinenumbers}
\begin{lstlisting}[style=promptstyle, caption={Code snippet of an additional discovered kernel with strong transferability. This kernel combines an RQ component with a B\'ezier-projected polynomial component.}, label={fig:discovered_code_bezier}]
def forward(self, x1, x2, diag=False, **params):
    if x1.dim() == 1: x1 = x1.unsqueeze(-1)
    if x2.dim() == 1: x2 = x2.unsqueeze(-1)
    
    # Rational Quadratic Transform
    x_centered_RQ1 = x1 / self.lengthscale_RQ
    x_centered_RQ2 = x2 / self.lengthscale_RQ
    x_scaled_RQ1 = x_centered_RQ1 / math.sqrt(self.D)
    x_scaled_RQ2 = x_centered_RQ2 / math.sqrt(self.D)
    
    # Bezier Path Projection
    t = self.harmonic_alpha
    path_bezier = sum((1-t)**(self.control_point_count-i) * t**i * self.bezier_control_points[..., i, :] 
                      for i in range(self.control_point_count))
    x_path_BZ1 = x1 / self.lengthscale_BZ
    x_path_BZ2 = x2 / self.lengthscale_BZ
    x_proj_BZ1 = self._inv_stereographic((x_path_BZ1 / x_path_BZ1.norm(dim=-1, keepdim=True) + path_bezier) / 2)
    x_proj_BZ2 = self._inv_stereographic((x_path_BZ2 / x_path_BZ2.norm(dim=-1, keepdim=True) + path_bezier) / 2)
    x_scaled_BZ1 = x_proj_BZ1 / math.sqrt(self.D)
    x_scaled_BZ2 = x_proj_BZ2 / math.sqrt(self.D) 
    
    # Rational Quadratic Kernel
    r_RQ2 = torch.cdist(x_scaled_RQ1, x_scaled_RQ2, p=2).pow(2)
    k_RQ = (1 + r_RQ2 / (2 * self.alpha_RQ)).pow(-self.alpha_RQ)
    
    # Bezier-Projected Polynomial Kernel
    k_BZ = (x_proj_BZ1 @ x_proj_BZ2.transpose(-1, -2) + 1).pow(3)
    
    # Hybrid Kernel Covariance
    k = k_RQ + self.harmonic_alpha * k_BZ
    
    if diag:
        return k.diagonal(dim1=-2, dim2=-1)
    return k
\end{lstlisting}
\end{nolinenumbers}

\clearpage
\vspace*{0pt}
\begin{figure}[t]
\begin{nolinenumbers}
\begin{lstlisting}[style=promptstyle, caption={Code snippet of another additional discovered kernel with strong transferability. This kernel combines RQ, spherical, Mat\'ern-5/2, and warped periodic-like components.}, label={fig:discovered_code_cosmic}]
def forward(self, x1, x2, diag=False, **params):
    if x1.dim() == 1: x1 = x1.unsqueeze(-1)
    if x2.dim() == 1: x2 = x2.unsqueeze(-1)

    # Rational Quadratic component
    x1_rq = x1 / self.lengthscale
    x2_rq = x2 / self.lengthscale
    r_rq2 = torch.cdist(x1_rq, x2_rq, p=2).pow(2)
    k_rq = (1 + r_rq2 / (2 * self.alpha)).pow(-self.alpha)

    # Spherical component
    x1_sphere = self._project_to_sphere(x1 / self.lengthscale_sphere)
    x2_sphere = self._project_to_sphere(x2 / self.lengthscale_sphere)
    coeffs = self.poly_coeffs_sphere
    phi1 = torch.cat([x1_sphere * coeffs[1].sqrt(), coeffs[0].sqrt() * torch.ones_like(x1_sphere[..., :1])], dim=-1)
    phi2 = torch.cat([x2_sphere * coeffs[1].sqrt(), coeffs[0].sqrt() * torch.ones_like(x2_sphere[..., :1])], dim=-1)
    k_sphere = phi1 @ phi2.transpose(-1, -2)

    # Matern-5/2 component
    x1_m = x1 / self.lengthscale_matern52
    x2_m = x2 / self.lengthscale_matern52
    r_m = torch.cdist(x1_m, x2_m, p=2).clamp(min=1e-15)
    k_m = (1 + math.sqrt(5) * r_m + 5 / 3 * r_m.pow(2)) * torch.exp(-math.sqrt(5) * r_m)

    # Cosmic Warping component
    x1_cosmic = torch.tanh(x1 / self.radius_cosmic)
    x2_cosmic = torch.tanh(x2 / self.radius_cosmic)
    cosmic_phase1 = x1_cosmic.norm(dim=-1, keepdim=True)
    cosmic_phase2 = x2_cosmic.norm(dim=-1, keepdim=True)
    k_cosmic = torch.cos(2 * math.pi * self.alpha_cosmic * torch.cdist(cosmic_phase1, cosmic_phase2, p=2))

    # Combined Kernel
    covar = k_rq + (k_sphere * k_m) + k_cosmic

    if diag:
        return covar.diagonal(dim1=-2, dim2=-1)
    return covar
\end{lstlisting}
\end{nolinenumbers}
\end{figure}

\clearpage


\end{document}